# Pattern-and-root inflectional morphology: the Arabic broken plural

Alexis Amid Neme [1] - Éric Laporte [2]

*Abstract*

*We present a substantially implemented model of description of the inflectional morphology of Arabic nouns, with special attention to the management of dictionaries and other language resources by Arabic-speaking linguists. Our model includes broken plurals (BPs), i.e. plurals formed by modifying the stem.*

*It is based on the traditional notions of root and pattern of Semitic morphology. However, as compared to traditional Arabic morphology, it keeps the formal description of inflection separate from that of derivation and semantics. As traditional Arabic dictionaries, the updatable dictionary is structured in lexical entries for lemmas, and the reference spelling is fully diacritized. In our model, morphological analysis of Arabic text is performed directly with a dictionary of words and without morphophonological rules.*

*Our taxonomy for noun inflection is simple, orderly and detailed. We simplify the taxonomy of singular patterns by specifying vowel quantity as* v *or* vv*, and ignoring vowel quality. Root alternations and orthographical variations are encoded independently from patterns and in a factual way, without deep roots or morphophonological or orthographical rules. Nouns with a triliteral BP are classified according to 22 patterns subdivided into 90 classes, and nouns with a quadriliteral BP according to 3 patterns subdivided into 70 classes. These 160 classes become 300 inflectional classes when we take into account inflectional variations that affect only the singular.*

*We provide a straightforward encoding scheme that we applied to 3 200 entries of BP nouns.*

## 1. Objective

We present a model of description of the inflectional morphology of Arabic nouns. Our purpose is to generate comprehensive dictionaries for Arabic natural language processing (NLP), and to equip them with easy procedures of manual, computer-aided updating. No such dictionary is currently available for Arabic NLP (cf. Section 2.4). Noun inflection is a crucial part of the inflectional system of Arabic: it regards a large part of the lexicon and ‘nouns turn out to be far more complex than verbs’ (Altantawy *et al*., 2010:851).[3]

Our approach, inspired from Neme’s work on verbs (2011), is to generate plurals from fully diacritized singular forms. The input of the system is a noun lemma with an inflectional code. The output is a list of inflected forms with their morpho-syntactic features. We take fully diacritized spelling as reference, and we deal with partially diacritized or undiacritized spelling through the concept of optional information.

We focus on broken plurals (BPs), defined as those Arabic plurals formed by modifying the stem, as in *Euqodap* ‘knot’ vs. *Euqad* ‘knots’. BPs contrast with suffixal plurals, which are formed by substituting suffixes, as in *Halaqap* ‘ring’ vs. *HalaqaAt* ‘rings’. A large proportion of nouns, e.g. most nouns of concrete objects and animals and many technical terms, have only a BP. ‘For the lexicon as a whole, then, broken plural formation is by far the norm rather than the exception’ (McCarthy, Prince, 1990:213).

In this paper, examples displayed in the Latin alphabet are transliterated according to Buckwalter-Neme (BN) code, a variant (Neme, 2011, p. 6, note 4) of Tim Buckwalter’s transliteration that avoids the use of special characters.[4] The diacritics for short vowels are noted *a, u, i*. A position between two basic letters without any vowel is noted *o*, as in *Euqodap* [ʕuqdap]. In other words, *o* does not note the [o] vowel, but is a silent diacritic: when it is noted, it rules out the hypothesis of a non-scripted short vowel. This transliteration system is entirely based on the digital encoding of text, as defined by the Unicode standard, and does not necessarily reflect its graphic display on the screen (e.g. ligatures) nor its pronunciation.

[1] LIGM, Université Paris-Est - alexis.neme@gmail.com
[2] LIGM, Université Paris-Est; DLL, Universidade Federal do Espírito Santo - eric.laporte@univ-paris-est.fr
[3] We thank Tim Buckwalter for helpful comments and discussions on an earlier version of this article.
[4] In this transliteration, upper-case and lower-case letters, e.g. *E* and *e*, denote distinct, independent consonants : ء, c; آ, C; أ, O; ؤ, W; إ, I; ئ, e; ا, A; ب, b; ة, p; ت, t; ث, v; ج, j; ح, H; خ, x; د, d; ذ, J; ر, r; ز, z; س, s; ش, M; ص, S; ض, D; ط, T; ظ, Z; ع, E; غ, g; ف, f; ق, q; ك, k; ل, l; م, m; ن, n; ه, h; و, w; ى, Y; ي, y; ◌ً , F; ◌ٌ , N; ◌ٍ, K; ◌َ, a; ◌ُ , u; ◌ِ , i; ◌ّ , G; ◌ْ, o.  The BN transliteration is implemented in the Unitex

## 2. Previous work

### 2.1. Root-and-pattern morphology

Among the possible formal representations of Arabic morphology, root-and-pattern morphology is a natural representation, as well as for other Semitic languages. It is so widely used that this model is also known as 'Semitic morphology'. A (surface) **root** is a morphemic abstraction, a sequence of letters, which can only be consonants or long vowels,[5] like *Eqd*, where *E* notes the pharyngeal or epiglottal consonant [ʕ], or *swr*, where *w* notes a long vowel in certain conditions. A **pattern** is a template of characters surrounding the slots for the root letters. These slots are shown in the pattern by indices, like in *1u2a3*. Between and around the slots, patterns contain short vowels, and sometimes consonants or long vowels. Once affixes are stripped off the surface form of a word, the remaining stem is analysed as the 'interdigitation' (Beesley, 1996) of a root with a pattern. For example, the stems *Euqodap* 'knot' and the BP *Euqad* 'knots' are represented by the root *Eqd* and, respectively, by the singular pattern *1u2o3ap* and BP pattern *1u2a3* :

| | | | |
|---|---|---|---|
| Stem | Euqodap | Euqad | عُقْـدَة عُقَـد |
| Root | E q d | E q d | |
| Pattern | 1u2o3ap | 1u2a3 | |

A root is usually stable across all the forms in a lexical item; grammatical distinctions between these forms correspond to different patterns. Thus, lexical items are classified in biliteral, triliteral, quadriliteral, quinqueliteral depending on the number of letters in their root. The general principles of root-and-pattern morphology are ubiquitous in the Arabic-speaking world and are taught in school. This representation is well established in Arabic morphology and seems well founded.[6]

There is a parallel between this model and Arabic script. Arabic script distinguishes 'basic letters', which are obligatorily written, and diacritics, which are usually omitted. All basic letters are consonants or long vowels, just as all root letters also are; roots are written with basic letters only. This is an additional reason why root-and-pattern morphology is so intuitive for users of Arabic script. Between and around the slots, patterns comprise diacritics, and sometimes basic letters.

The slots for root letters in a pattern are traditionally noted by the consonants *f, E, l, l,* instead of the digits *1, 2, 3, 4*. For instance, *1u2o3ap* and *1u2a3* are noted *fuEolap* and *fuEal* (فُـعْلَـة ، فُـعَل). This makes the representation of the pattern pronounceable, and thus easier to remember. We adopted this convention and adjusted it in several ways. We modified the consonant for the 4th slot, so as to have four different consonants *f, E, l, b*. When we script patterns in Buckwalter transliteration, we type these consonants in upper case: *F, E, L, B*, so that the slots are visually salient: *FuEoLap* and *FuEaL*. We note the long vowels *aa ii uu* instead of *aA iy uw*, which would be the fully diacritized BN transliteration. With this convention, adopted by several authors, the slots for the root consonants are easier to identify visually. They appear in capitals, while most other letters in patterns appear in lower case. When *aA* is written in BN transliteration, the upper case letter tends to confuse the recognition of the slots.

---

system (Paumier, 2002).

[5] As a simplification, we introduce here the surface root corresponding to a set of actually pronounced segments, and not the underlying root postulated by traditional Arabic grammar and by generative grammar.

[6] Prosodic morphology uses a close variant of this model (McCarthy, 1981) in which a pattern such as *1i2a3* is replaced by two abstractions: a 'CV skeleton' for the position of vowels, here *1v2v3*, and a 'melody' for their values, here *ia*. This variant is used in some implementations (Kiraz, 1994). We use the traditional form of patterns, which is simpler (Smrž, 2007:33) and more usual to Arabic speakers.

### 2.2. Traditional morphology

A large part of traditional Arabic morphology (TM), including the description of BPs, dates back to Sibawayh, a grammarian of the VIII$^{th}$ century (Sibawayh, ed. Haarun, 1977). Since then, his representation has been generally approved and transmitted by grammarians without major improvements. It is largely used at school in Arab countries.

This traditional view describes how BPs are produced from singular nouns. The path from a singular form to a BP passes through a root. The essential steps in this operation are:
- analysing the singular into a root and an existing singular pattern, e.g. *Euqodap* 'knot' = *[Eqd* & *FuEoLap]*,
- selecting a BP pattern, here *FiEaL*,
- combining the root with the BP pattern.

In the first step, we shift from a surface form to the root and pattern level; then, we shift back to surface. The steps listed above present four technical obstacles.

- The analysis of an Arabic word into a root and a pattern is not a deterministic operation and can a priori produce several results (cf. Section 4.1), even after discarding those results that violate any constraints about roots or patterns.

- TM's notions of root and pattern are not exactly the surface root and pattern introduced above, but a 'deep' root, e.g., in the case of *baAb* 'door' باب , *bwb* instead of *bAb*, and a 'deep' pattern. Rules modify these underlying forms to produce surface forms. Thus, the path from a singular form to a BP, in fact, passes through a deep root. To find the deep root, the rules have to be 'unapplied', i.e. applied regressively;[7] then, to generate the BP form, the same rules are applied back in the normal way.

- The BP pattern is generally unpredictable from the singular pattern.

- Once the root is combined with the BP pattern, rules apply and modify the deep forms.

Reliable dictionaries (Abdel-Nour, 2006) and excellent inventories of classes and nouns (Tarabay 2003) can be found. Sure, numerous entries in Tarabay are disused in Modern Standard Arabic, and some classes are missing, for example the human nouns with the *FaEaaLiBap* pattern in the BP, as *barobariyG* 'barbar' بربريّ or *malaAk* 'angel' ملاك . But the system is essentially unchanged since Sibawayh, and has incorporated loanwords harmoniously.

The TM model of BPs is precise enough to define taxonomies: two nouns are assigned the same class if they produce their BP in the same way. However, TM does not explicitly enumerate classes. The notion of taxonomy is also naturally connected with that of codes: two nouns belong to the same class if they are assigned the same BP codes. TM produces BPs from singular nouns through two 'codes': the first is either the singular pattern (*FiEoLap* in the example above) or the deep root (*Eqd*), and the second is the BP pattern (*FiEaL*).

Since Sibawayh, most lexicologists and linguists have contributed in the form of comments, rather than revisions. The accumulated comments tend to make the model seem more complex, not to simplify it. Among modern linguists, those who have adopted the root-and-pattern model have rarely questioned historical authors and practices either.

TM's model of BPs is complex. Tarabay's (2003) book about plural in Arabic, which is almost entirely dedicated to BPs, has 470 pages on 2 columns, plus 100 pages of glossaries representing more than 12 000 entries (not exhaustive, common words are lacking). BPs in themselves give an 'initial impression of chaos' (McCarthy, 1983:292) and are 'highly allomorphic' (Soudi *et al.*, 2002); grammatical and lexical traditions and practices along centuries do not give the impression of an effort towards a simpler and more orderly taxonomy, with fewer classes. Arabic specialists disagree about the deep root of some nouns, e.g. *xanoziyr* 'pig' خنزير is indexed under the roots *xnzr* and *xzr* in Ibn Manzur (1290) and under the root *xzr* in Al-Fairuzabadi (c. 1400). Descriptions of rules are often scattered in reference books, and their conditions of application are not formalized and not always fully specified. In a typical example, Tarabay (2003:92, footnote 2) mentions a metathesis rule that substitutes *o<cons>i<cons>* by *i<cons>o<cons>*, as in the underlying form **OaxoMiMap* 'vermin' أخشِشَة (the '*' symbol signals a reconstructed, not directly observed form) which takes the form **OaxiMoMap* أخِششَة., which in turn is correctly spelt as *OaxiMGap* أخِشَّة , where the *G* diacritic notes the gemination of the preceding consonant. She words the conditions of application as follows: '[The nouns] that pluralize only with the *OaFoEiLap* pattern, that have the *FaEaAL* pattern in the singular and that have identical 2nd and 3rd root letters, apply an *i* shift which is substituted by *o*.' In this footnote, 'pluralize only' means that the noun does not have another BP: if it has a suffixal plural, the rule can apply. Thus, the conditions of applications of this rule are and not fully specified.[8] There are dozens of such rules. Their order of application matters for their final output, but it is not systematically specified. Good traditional dictionaries explicitly provide BPs in surface form, bypassing the pattern and the rules.

[7] In case of doubt, lexicons provide the deep root directly.

The number of classes in a BP taxonomy measures the complexity of the BP system. Since TM does not count classes, let us compute estimations from numbers of patterns. Tarabay (2003) distinguishes 56 BP patterns. This number can be viewed as a measure of the complexity of BP: 'The defining characteristic of fixed-pattern morphology is that consistency in such systems is found not in a consistent proportion or relationship between two forms (a base and a derivative, an input and an output) but in a consistent pattern (of syllable structure and vocalism) imposed on all derived forms of a particular class regardless of the form of the source word' (Ratcliffe, 2001:153). However, the number of BP patterns underestimates the complexity of deducing a BP from a singular, because it overlooks the problem of finding the root. We should then take into account the number of singular patterns. The BP pattern is unpredictable from a given singular pattern, and vice versa, but not all singular pattern/BP pattern pairs are represented in the lexicon. Estimates of the number of singular pattern/BP pattern pairs vary from 105 (Murtonen 1964, survey based on the dictionary of Lane 1893) down to 55 (Soudi *et al*., 2002, citing Levy 1971, based on Wehr 1960) or 44 (El-Dahdah, 2002), but they are limited to the common types. Again, the number of pattern pairs does not take into account the additional complexity brought about by morphological variations. Such variations affect the consonants *w*, *y* and [ʔ] (the glottal stop), and forms with reduplicated or geminated consonants. Tarabay (2003) dedicates 30 pages to the latter type of variations. We estimate that her inventory is equivalent to more than 2 000 classes.

For TM, the description of BPs is required to be consistent with other constraints. For example, roots are also used for the practical purpose of indexing dictionaries. *'The lexical root provides a semantic field within which actual vocabulary items can be located'* (Ryding, 2005:677). Derived nouns such as *miEowal* 'mattock' معول are listed in dictionaries under the root of their base, here *Ewl*, a root that also occurs in words meaning 'howl', 'raise (a family)', 'rely on'... Therefore, the consonants of derivational prefixes, here *m*, are not analysed as being part of the root, even when they are common to the singular and BP of the derived noun, as is the case for this noun.

In a similar vein, the roots and patterns relevant for inflectional morphology are also 'reused' for semantic description. *'A root is a relatively invariable discontinuous bound morpheme, (...) which has a lexical meaning'* (Ryding, 2005:47). TM associates some patterns with semantic features, e.g. the *miFoEaL* pattern with the notion of instrument, as in *miEowal* 'mattock'. However, such associations are never perfectly regular. The *miFoEaL* pattern could not be used as a semantic label for instrument nouns. Some instrument nouns do not have it, e.g. *qalam* 'pen' قلم . The broken plural of *miEowal* 'mattock' معول itself, *maEaAwil* 'mattocks' معاول , is still an instrument noun, and has another pattern.

TM also integrates inflection with derivational morphology, which also involves roots and patterns. When a word is the output of a derivational process and the input of an inflectional process, as *miEowal* 'mattock', it is traditionally implied that its root-and-pattern analysis is the same with respect with the two morphological processes.

[8] Probably because it is relatively intuitive for Arabic speakers: *o<cons>i<cons>* sequences are rare in Arabic, and where they are expected, *i<cons>o<cons>* sequences are often observed.

Thus, notions relevant to production of BPs from singular nouns are reused for three other purposes: dictionary indexing, semantic description or derivational morphology. This integration makes sense in a context of Arabic teaching, in that it facilitates memorization. However, if we consider each of these four objectives separately, the reuse may lead to conflicting constraints, if the best definition of roots and patterns for the different purposes do not coincide exactly, as in the examples above. In addition, this integration makes the assignment of a word to a BP class depend on semantic and derivational information, and not only on inflectional morphology.

Summing up, the TM's account of BPs produces the correct forms, it has been tested and validated over centuries, and it is familiar to the Arabic speakers that are likely to encode and update lexical resources. Dictionaries have a readable layout and provide reliable information. However, there might be room for simplification:
- of the taxonomy,
- of the morphophonological rules,
- of the procedure of assignment of a noun to a class.

### 2.3. BP in generative grammar

Generative grammar gives several formal models of BP generation, some of them well documented, taking into account large portions of the Arabic lexicon, and based on interesting analyses. McCarthy & Prince (1990) propose a computation of BP stem from singular stem, a 'rule for forming the broken plural' (p. 263); Kihm (2006) formalizes other rules in a rival trend within generative grammar.

As compared to traditional morphology, these models hypothesize underlying forms and rules for surface realisation too, but they endeavour to lower the number of inflectional classes for BP. McCarthy & Prince (1990:210 and 217) view Wright's (1971) account of BP, with 31 plural types, corresponding to 11 singular types, as a 'poorly understood or perhaps even chaotic process', and they try to 'substantiate the informal notion that a single pattern unites all the classes grouped under the iambic rubric'. The price for reducing this 'apparent complexity' are more abstract underlying forms, i.e. more distance between underlying forms and surface forms, and therefore a more complex system of rules. The rules perform, for example, metathesis, after Levy (1971), and glide realisation, after TM and Brame (1970). The complexity of the systems comes from relations between rules, such as order of application, and from the existence of exceptions to them.

In conformity with the generative paradigm, these authors assume that the underlying roots exist in native speakers' minds and are activated during the production of BPs. We are not committed to this assumption, for lack of evidence; in addition, when several underlying roots are a priori possible, as in *qabow/Oaqobiyap* (see Section 3.3), we lack evidence about whether hypothetical underlying roots would be identical or different in respective speakers' minds. Our approach focuses on verifiable facts as much as possible.

The generativist models are not directly exploitable for computational purposes, for two reasons:

- The rules are only partially specified. McCarthy and Prince's (1990) rules rely on a metathesis (Levy, 1971) observed in *OakotaAf* for **kataAf* 'shoulders' أكتاف , but they leave undefined the conditions of application of the metathesis, not because they are easy to describe, but because they are 'not wonderfully transparent'. Instead of this metathesis, Kihm (2006:83) uses an 'augment of obscure origin', but does not specify the conditions of its insertion either. He also sketches rules according to which the 2$^{nd}$ root letter does not count as such when it is a glide, and another that integrates into the root some inflectional affixes of the indefinite singular during the generation of the BP (p. 86), but he does not explain in which conditions. As for the lexical information required to generate BPs, he 'leave[s] the precise formalization of this information to future work' (p. 81). Similarly, McCarthy & Prince do not enter into details to the point that they would tell how many inflectional classes for BP should be distinguished with their model.

- Nouns showing exceptional behaviour are mentioned, but not dealt with in the models. For example, McCarthy & Prince's (1990:273-274) rules with left-to-right association give the correct BP in many quinqueliteral nouns, but they do not propose any device for exceptions, since generative grammar is not committed to describing lexical items beyond those that 'reflect a regular grammatical process of the language' (p. 267). Generative grammar aims to model a specifically linguistic mental process, and is traditionally not interested in general-purpose mnemonic processes that are supposed to handle exceptions when they are not too numerous. This is an important difference with our objectives, since a comprehensive morpho-syntactic lexicon is required to deal with all cases.

Anyway, the generative models of BP, even incompletely specified, seem already too complex to be the best choice for our practical objective of a system easy to update. Complex relations between rules, such as order of application, and the existence of exceptions to them, obfuscate these systems.

In addition, this additional complexity of the rules (as compared to TM) does not always contribute to simplify the taxonomy of BPs. For example, McCarthy & Prince (1990) predict the quantity of last *i* in quadriliteral BP patterns when the first syllable of singular is bimoraic in the generative sense. This allows for merging the *FaEaaLiB* and *FaEaaLiiB* patterns for some nouns. With reference to the goals of generative grammar, such a prediction makes sense, since it models a linguistic process by a rule which is assumed to comply with a universal format. However, if we now take in mind our goal of simplifying the encoding of lexical items, the prediction tends to complicate the generation of the BP, without lowering the number of patterns, since *FaEaaLiB* and *FaEaaLiiB* must still be distinguished for the BP nouns whose first syllable is not bimoraic.

Kihm (2006:81) claims that his model simplifies dramatically the taxonomy of BPs: 'such a wild variety of forms actually results from one process and from the interplay of a few well defined factors', namely the timbre of an element inserted between the 2$^{nd}$ and 3$^{rd}$ root letters, which is chosen between *i*, *a* and *u*, and the category of the insertion: consonant or vowel. However, this claim overstates the simplicity of Kihm's taxonomy. In his model, the variety of forms also depends, for example, on the value of the vowel inserted between the 1$^{st}$ and 2$^{nd}$ root letters of the BP (p. 82).

### 2.4. Analysers and generators of Arabic inflected words

Because of the rich morphology of Arabic, NLP for this language requires dictionaries: 'we need to be able to relate irregular forms to their lexemes, and this can only be done with a lexicon' (Altantawy *et al.*, 2010:851). This need also applies to the statistic methods which are widely expoited almost without dictionaries for other inflectional languages: 'the need for incorporating linguistic knowledge is a major challenge in Arabic data-driven MT. Recent attempts to build data-driven systems to translate from and to Arabic have demonstrated that the complexity of word and syntatic structure in this language prompts the need for integrating some linguistic knowledge' (Zbib, Soudi, 2012:2).

Still, no comprehensive dictionaries equipped with easy procedures of updating are currently available for Arabic NLP. In the last 20 years, a number of computer systems for the morphological analysis and generation of Arabic words have been implemented. They can be classified into two approaches.

- The root/pattern/rule approach is based on traditional morphology. During analysis, a stem is analysed into a deep root and a deep pattern which are looked up among the roots and patterns stored in the system. The distance between deep level and surface level is covered with the aid of rules. This approach has a variant where patterns are closer to the surface, reducing the distance and simplifying the rules.

- The multi-stem approach seeks to avoid heavy computation during analysis. A stem is looked up among the stems stored in a dictionary. The term 'multi-stem' alludes to the fact that a lexical entry for a BP noun or a verb has at least two stems, e.g. *miEowal* 'mattock' معول and *maEaAwil* 'mattocks'. This approach has a variant in which the stems are generated from roots and patterns during a dictionary compilation phase.

<u>2.4.1. Beesley (1996, 2001)</u>

This system for Arabic inflection formalizes the traditional version of the root-and-pattern model and classifies in the root/pattern/rule approach. Its rules deal with root alternations, morphophonological alternations and spelling adjustments. They are encoded in the form of finite automata and compiled with the dictionary into a finite transducer. For morphological analysis, these rules are applied regressively, i.e. they take surface forms as input and they output deep forms.

The system has a medium lexical coverage: 4 930 roots producing 90 000 stems [9] (Beesley, 2001:7), and it includes BPs. The lexical data originate from work at ALPNET (Buckwalter, 1990).

This system faces several challenges. One of them is that of analysis speed: 'the finite-state transducers (FSTs) tend to become extremely large, causing a significant deterioration in response time' (Altantawy *et al.*, 2011:116). This was, by the way, the main motivation for devising the multi-stem approach.

A second problem is the complexity of the rules that produce surface forms from underlying forms. The deep roots are borrowed from traditional morphology. For example, *baAeiE* 'seller' بائع , with surface root *beE*, and *baAEap* 'sellers' باعة (cf example 79 below), with surface root *bAE*, are analysed with deep root *byE*, which requires that the rules change *y* into *e* in the singular and into *A* in the plural. Each difference between surface forms and deep forms increases the complexity of the rule system. This complexity does not bring about any identifiable benefit. Once the roots are output by the analyser, they are to be essentially used as morpheme labels: the deep root borrowed from traditional morphology is not better for that than, say, the surface root of the singular. This additional complexity is inherited from traditional morphology, where it is meant to contribute to the semantic indexing of dictionaries, and to the consistency between inflection and derivation (Section 2.2 above). A morphological analyser of Arabic does not need to take into account these constraints: semantic indexing has no relation with morphological analysis; nobody finds it necessary to integrate inflection and derivation, for example, in English, in spite of obvious regularities between derivational suffixes and inflectional properties. "Dictionary maintenance need not require a thorough knowledge of Arabic derivational morphology, which few native speakers learn" (Buckwalter, 2007:37). And the useless complexity induced by the deepness of the underlying level has a cost: the rules are encoded and updated manually, 'a tedious task that often influences the linguist to simplify the rules by postulating a rather surfacy lexical level' (Beesley, 1996:91).

A third problem with this system is that the model lacks the notion of inflectional class. Two nouns belong to the same inflectional class if they inflect in the same way, and in particular if they pluralize in the same way. In lexicology for language processing, this notion allows for devising a common process shared by all the entries of a class, making the complexity depend on the number of classes (typically a few hundred) rather than on the number of lexical entries (in the dozens of thousands). Take for example root alternations: the surface root of *baAeiE* 'seller' بائع is *beE* in the singular and *bAE* in the BP, whereas for *HaAeir* 'indecisive' حائر ,[10] it is *Her* in the singular and *Hyr* in the BP (cf example 78 below). Considering that there are no inflectional classes amounts to considering that both entries pluralize in the same way. This imposes to design and implement a **single** set of rules that outputs the correct alternation for both — and for **all** entries of all classes, in addition to the fact that for each entry, it should produce both the correct singular and the correct BP. In practice, this is a real challenge: 'Not surprisingly, to anyone who has studied Arabic, the rules controlling the realization of **w**, **y** and the hamza (the glottal stop)[11] are particularly complicated' (Beesley, 2001:5). Checking, correcting and updating such a set of rules are also heavy tasks: a typical rule affects several kinds of lexical entries, and there is no index of the entries or classes affected by each rule, or of the rules affecting each entry or class; the order of application of the rules is significant and must be decided and encoded. A separate, simpler set of rules for each class is more convenient to handle, even if at the cost of some redundancy between classes.

The solution adopted to specify BP patterns is diametrically opposed to the one for root alternations: patterns are manually specified separately for each root (Beesley, 2001:7), without sharing information at the level of inflectional classes.

---

[9] It is not measured as a number of entries because the formal model of the system does not include the notion of lexical entry.

[10] The BP system is essentially the same for nouns and for adjectives, except that BP is stylistically preferred for nouns, and suffixal plural for adjectives. We will exemplify some facts with adjectives.

[11] The consonants *w*, *y* and [ʔ] mentioned here are precisely those involved in root alternations.

The final shortcoming of this system is the format of the output of analysis, at the 'abstract lexical' level. It identifies the POS, root and pattern of the analysed words and their inflectional features, but not their lexical entries. Lexical entries of words are used to store, for example, their syntactic and semantic features, or, in the case of multilingual systems, an index to a lexical entry in another language. For example, *EawaAeil* 'families' is analysed by the system as a noun with root *Ewl* and pattern *FaEaAeiL*, and *maEaAwil* 'mattocks' as a noun with the same root and pattern *maFaAEiL*, but this is insufficient to identify lexical entries for them: since both words share the same root *Ewl*, nothing specifies whether one of them is the plural of *EaAeilap* 'family' or of *miEowal* 'mattock'. This is a difference with traditional dictionaries, which have a level for lexical entries in addition to the level for roots.

2.4.2. MAGEAD

The MAGEAD system (Habash, Rambow, 2006; Altantawy *et al*., 2010, 2011) is close to Beesley's (2001) in its design: 'We use "deep" morphemes throughout, i.e., our system includes both a model of roots, patterns, and morphophonemic/orthographic rules, and a complete functional account of morphology' (Altantawy *et al*., 2010:851); the rules are also compiled with the lexicon into a finite transducer. The lexicon is derived from Buckwalter's (Habash, Rambow, 2006:686; Altantawy *et al*., 2010:853) through Smrž's (2007). The project has an on-going part for nouns, including BPs (Altantawy *et al*., 2010).

MAGEAD improves upon Beesley (2001) in several ways. The notion of lexical entry is represented. The output of morphological analysis of a noun comprises sufficient information to identify a lexical entry in the same way for the singular and the plural (Altantawy *et al.*, 2010:853): for *mawaAziyn* 'balances', the lexical entry of the noun is identified by the root *wzn* and the '*noun-I-M-mi12A3-ma1A2iy3*' codes, which specify the part-of-speech, the non-human feature, the gender and the compatibility with patterns. This makes the results of morphological analysis more easily usable in other tools. The notion of inflectional class is adopted for patterns, but not for root alternations (Habash, Rambow, 2006:683): each lexical entry is assigned a code that identifies the patterns it admits, e.g. '*mi12A3-ma1A2iy3*' (Altantawy *et al*., 2010:853). There are 41 classes for verbs (Habash, Rambow, 2006:684). Thus, inflectional information is shared at class level, reducing redundancy between entries. This facilitates dictionary checking, update and extension, reducing the cost of management of the dictionary: when an error is detected in the patterns of a class, the correction of the error affects all the class; when a new class is found and encoded, it can be shared by all the future members of the class through a simple code assignment.

However, MAGEAD still faces the other problems that we mentioned above about Beesley (2001).

- The resources of MAGEAD-Express compile in 48 h, and the analysis of a verb takes 6.8 ms (Altantawy *et al.*, 2011:123).

- The analysis opts for deep roots, complexifying the computation of the root from the surface form.

- Root alternations are not taken into account in inflectional classes, but controlled by a single set of rules for all entries. Encoding such rules is a challenge: 'we also exclude all analyses involving non-triliteral roots and non-templatic word stems since we do not even attempt to handle them in the current version of our rules' (Altantawy *et al.*, 2010:856).

In addition, the lexical coverage is still limited. The lexical data are borrowed from Buckwalter (2002): 8 960 verbs (Altantawy *et al.*, 2011:122) and 32 000 nouns, including those with suffixal plural (Altantawy *et al.*, 2010:854), but the rules are compatible only with triliteral nouns: 'we are not evaluating our lexicon coverage (…) Our evaluation aims at measuring performance on words which are in our lexicon, not the lexicon itself. Future work will address the crucial issue of creating and evaluating a comprehensive lexicon' (Altantawy *et al.*, 2010:856).

2.4.3. Systems with root alternations encoded in patterns

The Elixir system (Smrž, 2007) has a medium lexical coverage and includes BP. The lexical data are adapted from Buckwalter (2002). It is slow, but could be quicker if implemented in another language than Haskell. The results include a representation of lexical entries, as in MAGEAD.

Elixir follows the root/pattern/rules approach, but, as compared to the systems described above, patterns are closer to the surface level. In case of root alternation, surface forms of root letters are specified in patterns. For example, *baAEap* 'sellers' is analysed with root *byE* and pattern *FaaLap*, whereas traditional morphology taught at school analyses it with root *byE* and deep pattern *FaEoLap*, with *A* as the surface realisation of the second deep root letter *y*. Traditional morphology represents patterns with a fixed number of slots, even in case of root alternations. Elixir's option of encoding root alternations in patterns is shared by Ryding (2005:149): *FaaLap, FuEaap*…(فَالة, فُعَاة...) This option simplifies the rules and their application, but introduces numerous new patterns, which look odd to Arabic speakers because traditional inflectional taxonomy is entirely based on deep patterns. This difference makes some of the Elixir patterns difficult to read and handle. In NLP companies, management of Arabic language resources tends to involve native Arabic speakers, because of their wider knowledge of the language.

The open-source Alkhalil morphological analyser [12] (Boudlal *et al.*, 2010) is used in various projects and won the first prize at a competition by the Arab League Educational, Cultural Scientific Organization (ALESCO) in 2010. We counted that Alkhalil's lexical resources cover 97% of the verb occurrences of a sample text, which is comparable to the coverage of Buckwalter (2002). The system includes BP. The patterns are scripted in Arabic. As in Beesley (2001), the output of the analyser does not identify lexical entries: nothing connects a noun in the BP to its singular. The general approach is close to that of Elixir, patterns are used in the same way, and the example of *baAEap* 'sellers' gets the same analysis.

Another difference with traditional morphology is that Alkhalil includes case and definiteness suffixes in the patterns. For example, in the noun *daAra* دَارَ 'home', Alkhalil assigns final -*a* to the pattern *FaaLa* فَالَ, whereas for traditional morphology, the stem is *daAr*, with root *dwr* and deep pattern *FaEaL* فعل (with *A* as the surface realisation of the second root letter *w*), and -*a* is an inflectional suffix of the accusative case and the construct-state definiteness. Traditional morphology has a systematic delimitation between stem and such suffixes; these suffixes have very little variation depending on lexical entries; most analysers comply with this distinction and exclude the suffixes from the pattern. The Alkhalil option introduces numerous such new patterns which are alien to familiar pattern taxonomy.

2.4.4. The multi-stem approach

Buckwalter's (2002) open source morphological analyser of Arabic, BAMA, is a well-known example of the multi-stem approach. It is slow, but could be quicker if implemented in another language than Perl. It has a medium lexical coverage: approximately 32 000 nouns and 9 000 verbs. The lexical data originate probably from work at ALPNET, as can be seen by the common morpheme labels (Buckwalter, 1990:3-5). All stems are stored in the resources, including most spelling variants, bypassing almost all morphophonological rules. This option simplifies dramatically the lookup algorithm. 'The BAMA uses a concatenative lexicon-driven approach where morphotactics and orthographic adjustment rules are partially applied into the lexicon itself instead of being specified in terms of general rules that interact to realize the output' (Buckwalter, 2002). Thus, 9 stems are stored for the verb *qara>a* 'read' قرأ (in Buckwalter transliteration), due to the orthographic variants of the 3rd root letter, here [ʔ], determined by the presence of an inflectional suffix or of an agglutinated pronoun. The form *qora>* appears in 3 items, with different compatibility codes:

[12] http://sourceforge.net/projects/alkhalil/

| Stem | Compatibility code | Stem | Compatibility code |
|---|---|---|---|
| *qara>* | PV-> | *qora\|* | IV-\| |
| *qara\|* | PV-\| | *qora&* | IV_wn |
| *qara&* | PV_w | *qora}* | IV_yn |
| *qora>* | IV | *qora>* | IV_Pass |
| *qora>* | IV_wn | | |

The information provided in morpheme labels includes the part of speech, the voice and aspect of verbs, and other relevant information.

Independent work by Soudi *et al*. (2002) shares the same design: ‘Such an approach dispenses with truncating/deleting rules and other complex rules that are required to account for the highly allomorphic broken plural system’ (Soudi *et al*., 2002). The main difference is that in case of purely orthographic variations, variants of stems are not stored in the lexicon, but the paper does not explain how they are recognised.

To date, the systems implementing the multi-stem approach have several common shortcomings. The multi-stem model lacks the notion of inflectional class: stems are manually specified separately for each root. For example, if a verb conjugates like *qara>a*, its 9 stems are listed independently of those of *qara>a*, without sharing information at the level of inflectional classes.

In addition, for a BP noun without root alternations, such as *EaAeilap* ‘family’ عائلة, *EawaAeil* ‘families’ عوائل , the stems stored in the lexicon include redundancy. The same root appears in each stem. Duplicated manual encoding of the same piece of information leads to errors. This flaw is connected to the preceding: multi-stem systems do not encode regularities.

Both have practical consequences. Human operations required to encode, check, correct and update the dictionaries are unnecessarily repetitive and costly. Fallback procedures for words not found in the dictionary are difficult to devise.

2.4.5. Neme (2011)

Neme (2011) describes a morphological analyser for Arabic verbs with a comprehensive lexical coverage: 15 400 verbs. The dictionary compiles in 2 mn and the analysis of a verb takes 0.5 ms on a 2009 Windows laptop,[13] outperforming MAGEAD-Express (cf. Section 2.4.2).

This system shows a concern with the comfort and efficiency of human encoding, checking and update of dictionaries. NLP companies need easy procedures for dictionary management, because most projects involve a specific domain with a particular vocabulary, and terminology evolves constantly; in addition, dialects show lexical differences, which are relevant to speech processing if not for written text processing; finally, the main advantage of dictionary-based analysers is that they provide a way of controlling the evolution of their accuracy by updating the dictionaries. None of the other authors surveyed above mentions the objective of facilitating manual dictionary management, and we reported the weak points of their analysers in this regard. Neme (2011) identifies the problem as belonging not only to computation and morphology, but also to NLP dictionary management, and considers language resources as the key point, as Huh & Laporte (2005). His dictionaries are constructed and managed with the dictionary tools of the open-source Unitex system (Paumier, 2002).

All forms are stored in the resources, including spelling variants; roots and patterns are handled at surface level. The main difference with previous multi-stem systems is that the full-form dictionary is automatically precompiled from another dictionary, which is specifically dedicated to manual construction, check and update. The dictionary is compiled by finite transducers that combine roots, patterns and inflectional suffixes. Each of the 480 inflectional classes is assigned one of the transducers, which ensures that the management of classes is mutually independent. The encoding of a new verb amounts to assigning it an inflectional code. Thus, the redundancy problems of the mainstream multi-stem approach are solved.

[13] Memory: 16 GB DDR3 1600 MHz; hard disks: 750 GB (7 200 rpm, Hybrid 4 GB Serial ATA) and 1TB (5 400 rpm, Serial ATA).

Pattern taxonomy is kept simple and close to that taught in school to Arabic speakers, by maintaining it separate from the encoding of root alternations and of tense, person, gender and number suffixes. This keeps codes readable and facilitates the encoding, improving upon the pattern labels of Smrž (2007) and Boudlal *et al*. (2010).

Such technology reduces the computational skills required for the linguistic part of dictionary management: these skills shift from software development to software use. Such a shift opens the perspective that Arabic language resources can be managed directly by native Arabic linguists. In current practice, management of resources typically requires a high-wage specialist of computation and an Arabic informant: a configuration which is more costly and inserts an intermediary between the source of linguistic knowledge and the formalization.

The results with verbs incited us to undertake the encoding of the BP system on the same bases. We called our project Pattern-and-Root Inflectional Morphology (PRIM), inverting the traditional 'root-and-pattern' phrase, because we capitalized on traditions about patterns, rather than about roots, to make our taxonomy intuitive to Arabic speakers.

## 3. General organization of PRIM

We decided to take advantage of the validation of traditional morphology over centuries, and we took it as a basis for our computerized model of BPs, formalizing and simplifying it. We gave priority to this objective of simplification in order to make easier and more comfortable the manual part of the encoding of Arabic dictionaries. Consistency with semantic features or derivational analyses was only a secondary objective. The most successful projects of morpho-syntactic codification are usually those that focus, in practice, on manual descriptors' ease and comfort. They produce long-lasting morpho-syntactic dictionaries which are actually updated over time by linguists, as has been the case of the Dela dictionaries since the 1980s (Courtois, 1990; Daille *et al*., 2002).

### 3.1. Inflectional codes

Arabic grammarians usually display the analysis of a singular stem/BP stem pair, e.g. *Euqodap* 'knot'/*Euqad* 'knots', in the form of a compact formula:

```
(a)                Eqd FuEoLap FuEaL
```

where *Eqd* is the deep root, *FuEoLap* the singular pattern and *FuEaL* the BP pattern. By combining *Eqd* with *FuEoLap* and applying morpho-phonological and orthographical rules, one obtains the singular stem. The same operation with *Eqd* and *FuEaL* yields the BP stem.

Pattern pairs such as *FuEoLap*/*FuEaL* make up a taxonomy of BP noun entries, by crossing the two taxonomies based, respectively, on singular patterns and BP patterns. A given singular pattern is compatible with several BP patterns, but not with all, and vice-versa.

The PRIM format of a lexical entry is similar to (a), with the lemma in Arabic script and the codes in the Latin alphabet:

```
(b)                Euqodap,  $N3ap-f-FvEvL-FuEaL-123
```

In this entry, *Euqodap* is the lemma of the noun, which is the singular of the noun, stripped off of its case and definiteness suffix, and written in fully diacritized script. The remainder is the inflectional code provided by the dictionary. In this code, *FvEvL* and *FuEaL* are the PRIM counterparts of the two patterns *FuEoLap* and *FuEaL* in (a), and the root code *123* is comparable to the deep root *Eqd* in (a). Our encoding of nominal entries is also similar to that of verbal entries (Neme, 2011), with two patterns and a root code:

```
Euqodap,  $N3ap-f-FvEvL-FuEaL-123          / knot
kaAotib,  $N300-g-FvvEvL-FuEEaaL-123       / author, employee
kaAotib,  $N300-g-FvvEvL-FaEaLap-123       / employee
katiyobap,$N3ap-f-FvEvvL-FaEaaLiB-12h3     / brigade of soldiers
```

```
kitaAob,  $N300-m-FvEvvL-FuEuL-123           / book
ktb,      $V3-FaEaLa-yaFoEuLu-123            / write
Inktb,    $V3-IinoFaEaLa-yanoFaEiLu-123      / be written
tkAtb,    $V3-taFaaEaLa-yataFaaEaLu-123      / write each other

عِقْدَة,    $N3ap-f-FvEvL-FiEaL-123            / knot
كَاْتِب,    $N300-g-FvvEvL-FuEEaaL-123         / author, employee
كَاْتِب,    $N300-g-FvvEvL-FaEaLap-123         / employee
كَتِيْبَة,   $N3ap-f-FvEvvL-FaEaaLiB-12h3       / brigade of soldiers
كِتَاْب,    $N300-m-FvEvvL-FuEuL-123           / book
كتب,      $V3-FaEaLa-yaFoEuLu-123            / write
إنكتب,    $V3-IinoFaEaLa-yanoFaEiLu-123      / be written
تكاتب,    $V3-taFaaEaLa-yataFaaEaLu-123      / write each other
```

In verbal entries, the two patterns are for the perfect and the imperfect. Verb lemmas are encoded without diacritics; the diacritics are specified in the perfect pattern.

### 3.2. Special plurals

As a simplification, our model does not take into account the traditional marking of a few BP forms as 'plurals of paucity'. Sibawayh (VIII$^{th}$ century) states that in an older stage of Arabic, plural of paucity had been restricted to collections of 3 to 10 entities, and other plural forms to collections of more than 10; however, at his time, both constraints were commonly overlooked, and many nouns lacked a plural of paucity (Ferrando, 2002:5). Native speakers accept a 'non-paucity' BP after cardinal numbers from 3 to 10, even when the noun also has a plural of paucity:

عليه أن يختار ثلاثة ( أمكنة + أماكن )
*Ealayhi Ono yaxotaAr valaAvapu (Oamokinapin + OamaAkina)*
on him to choose three (places+pauc + places)
'He must choose three places'

عليه أن يختار أربعَ ( أيدي + أيادي)
*Ealayhi Ono yaxotaAr OarobaEa (Oayodin + OayaAdi)*
on him to choose four (hands+pauc + hands)
'He must choose four hands'

In addition, the delimitation of plural of paucity is fuzzy. Four BP patterns are associated to plurals of paucity, but they also generate non-paucity BPs. Grammars give examples of plurals of paucity, but never exhaustive inventories.

We do not mark 'plurals of plurals' either. Plurals of plurals in TM, as *OamaAkin* 'places', are supposedly obtained by morphologically pluralizing a BP, here *Oamokinap* 'places', which is re-pluralized on the same model as *zawobaEap* 'tornado'/*zawaAbiE* 'tornados'. In our model, *OamaAkin* 'places' is directly related to the singular *makaAn* 'place'.

As a rule, the PRIM taxonomy gives only one plural of a given lexical entry: when several plurals are observed, they are assigned to distinct entries, no matter whether they are equivalent or not, as in examples (86), (97) and (119). Neme (2011:7) discusses the same problem for verbs. When several entries generate identical singular forms, the Unitex system removes duplicates.

### 3.3. Interpretation of codes

The main 3 codes in a PRIM entry for a BP noun, as *FvEvL-FiEaL-123* in (b), correspond to 3 independent taxonomies which, crossed together, are sufficient to identify the generation of a broken plural.

The linguistic interpretation of these codes correspond to three conceptual steps in generating a BP from a lemma such as *Euqodap* 'knot': extract the surface root of the lemma, here *Eqd*; find out the surface root for the BP, which is unchanged here; and combine it with the BP pattern, which gives *Euqad* 'knots'.

The first step matches the singular-pattern code, here *FvEvL*, with *Euqodap*, to obtain *Eqd*:

| | | | |
|---|---|---|---|
| Stem | Euqodap 'knot' | qabow 'cave' | قَبْو |
| Singular-pattern code | FvEvL | FvEvL | |
| Surface root of singular | E q d | q b w | |

The second step applies root alternations [14] encoded in the root code, if any, as is the case with *12y*, the root code of *qabow* 'cave':

| | | |
|---|---|---|
| Surface root of singular | Eqd | qbw |
| Root code | 123 | 12y |
| Surface root of BP | Eqd | qby |

The third step combines the surface root with the BP pattern:

| | | | |
|---|---|---|---|
| Surface root of BP | E q d | q b y | |
| BP pattern | FuEaL | OaFoEiLap | |
| BP stem | Euqad 'knots' | Oaqobiyap 'caves' | أقبية |

Lemmas with a geminated consonant are a little more complex. In Arabic script, the *G* diacritic notes the gemination of the preceding consonant. For example, *MidGap* 'trouble' is to be read as if it were spelt **Midodap*. The silent diacritic *o*, which marks the absence of vowel (cf. Section 2), is not used when *G* is used. In this word, the singular-pattern code *FvEvL* implies that the geminated consonant corresponds to two slots in the root. The gemination is assigned to the root:

| | gloss | sg. stem | PRIM codes | sg. root | in Arabic |
|---|---|---|---|---|---|
| 1 | trouble | MidGap | FvEvL-FaEaaLiB-12h2 | Mdd | شدّة شدائد |
| 2 | luck | HaZG | FvEvL-FuEuuL-122 | HZZ | حظّ حظوظ |

In *sulGam* 'ladder', the geminated consonant corresponds to a single slot in the root, which is represented by a repeated letter in *FvEEvL*. The gemination is assigned to the pattern:

| | | | | | |
|---|---|---|---|---|---|
| 3 | ladder | sulGam | FvEEvL-FaEaaLiB-1223 | slm | سُلَّم سَلالِم |

The choice between the two analyses is determined by observing other forms and specified in the singular-pattern code.

In the Unitex implementation of PRIM, the three conceptual steps described above are performed simultaneously by inflectional transducers, as in Silberztein (1998). For example, in the transducer for inflectional class *N3ow-m-FvEvL-OaFoEiLap-12y*, which is the class of *qabow* 'cave', they are performed by formula *Oa1o2iy*, where *1* and *2* refer to the positions of root letters in the lemma, *y* is the value of the other root letter in the plural, and the remaining symbols correspond to the BP pattern; the *-ap* suffix in the pattern is specified in another part of the transducer, because it undergoes spelling variations in the presence of a clitic pronoun.

### 3.4. Encoding nouns

Encoding a noun consists of writing the stem of its lemma in fully diacritized form, and assigning it a code as in (b) (with the lemma in Arabic script), so as to generate the correct forms of the plural:[15]

```
(b)              Euqodap,  $N3ap-f-FvEvL-FuEaL-123
```

It is important that the stem is fully diacritized, since digits in inflectional transducers refer to the position of root letters. Each basic letter, except the last of the stem, is followed by a single diacritic, which is either a short vowel: *a*, *u*, *i*, or the void diacritic *o*. Thus, all root letters correspond to odd positions. The only exceptions are after a geminate consonant, which is transcribed as in example (3): the 3rd root letter, *m*, is in position 6.

[14] We term as 'root alternations' any changes in the surface value of root letters, as in *qabow* 'cave' قَبْو and *Oaqobiyap* 'caves' أقبية, or in the number of root letters, as in *TaAbiE* 'stamp' طابع and *TawaAbiE* 'stamps' طوابع (cf. Section 5).

[15] Computer aiding could be devised to assist encoders, but might have perverse effects, e.g. inciting them to systematically accept suggestions, even if they are inconsistent with previously encoded entries.

The choice of a code is not a deterministic process, because analysis in root and pattern is in general not deterministic (cf. (1)-(3) above, and Section 4.1). Traditional morphology provides rules for reducing indeterminacy. Our taxonomy complies with rules which are widely known by Arabic speakers: for example, triliteral roots take precedence over biliteral roots. However, we disregard rules that depend on scholarly or diachronic knowledge, when this reduces the number of classes or simplifies the task of assigning a class to a lexical item.

## 4. Conflation of patterns

In order to make the PRIM taxonomy of BPs simpler than the traditional one, we merged classes by conflating patterns without loss of information. We illustrate this in the following examples.

### 4.1. Singular patterns

The PRIM models substitutes singular-pattern codes, e.g. *FvEvL*, to the traditionally used singular patterns, e.g. *FiEoLap*. The PRIM singular-pattern codes are less numerous than singular patterns because they dispense with unnecessary information. Their only purpose is to be matched with lemmas, e.g. *Euqodap*, to obtain their surface roots, here *Eqd*:

| | |
|---|---|
| Stem | `Euqodap` 'knot' |
| Singular-pattern code | `FvEvL` |
| Surface root of singular | `E q d` |

The singular-pattern code cannot be dispensed of completely. Some nouns have more than three root consonants: the singular-pattern code *FvEvLvB*, matched with *diroham* 'dirham' درهم , extracts the root *drhm*. The difference between the two surface forms *Euqodap* and *diroham* would not be easy to tell without these codes.

Similarly, some noun lemmas have a long vowel, which is assigned either to the root or to the pattern. In *Miyomap* 'honour' شِيْمَة, the *iy* sequence [16] notes the long vowel [i:]; the *FvEvL* code implies that the root is *Mym*. The root letter *y* is realised as a long vowel. In contrast, in *sabiyol* 'road' سَبِيل , the *FvEvvL* code points to the root *sbl*. The long vowel belongs to the pattern.

Thus, simplified singular patterns such as *FvEvL*, *FvEvLvB*, *FvEEvL* or *FvEvvL* specify the number of root letters, the position of pattern-assigned long vowels, and the position of pattern-assigned geminations of root letters. They are sufficient to deduce the singular root.

Representing *o*, the silent diacritic, by *v*, a symbol for a short vowel, might seem paradoxical, but it is natural to Arabic speakers.

#### 4.1.1. Omission of vowel quality

The quality of the vowels is not specified because it is not necessary. This reduces the number of classes in the singular-pattern taxonomy, without loss of generative power. In the following examples, 6 singular patterns distinguished by TM are conflated into a single code in the PRIM model:

[16] In Arabic script, the letters *y* and *w* code the semivowels [j w] or the long vowels [i: u:], depending on context. When *y* is preceded by *a* or *u*, it codes [j]; when *w* is preceded by *a* or *i*, it codes [w]. The long vowels [i: u:] are coded *iy* and *uw*. This system codes alternations between [i: u:] and [j w]. The silent diacritic *o*, which notes the absence of vowel between two basic letters (cf. Section 2), is usually omitted after long vowels (*iy*, *uw*, *aA*), even when writers intend to fully diacritize their text. However, the PRIM model requires that it be present in lemmas, so that the convention given in Section 3.4 is respected, and roots with semivowels do not require separate classes. For the sake of consistency, from here on, this diacritic will be explicitly scripted in our examples.

| | gloss | sing. | plural | TM patterns | PRIM codes | Arabic |
|---|---|---|---|---|---|---|
| 4 | spirit | nafos | nufuwos | FaEoL-FuEuuL | FvEvL-FuEuuL-123 | نفس نفوس |
| 5 | luck | HaZG | HuZuwoZ | FaEo**L**-FuEuuL | FvEv**L**-FuEuuL-122 | حظّ حظوظ |
| 6 | stem | jiJoE | juJuwoE | FiEoL-FuEuuL | FvEvL-FuEuuL-123 | جَذع جذوع |
| 7 | load | Humol | Humuwolap | FuEoL-FuEuuLap | FvEvL-FuEuuLap-123 | حُمْل حمولة |
| 8 | mountain | jabal | jibaAol | FaEaL-FiEaaL | FvEvL-FiEaaL-123 | جَبل جبال |
| 9 | shoulder | katif | OakotaAof | FaEiL-OaFoEaaL | FvEvL-OaFoEaaL-123 | كتف أكتاف |
| 10 | man | rajul | rijaAol | FaEuL-FiEaaL | FvEvL-FiEaaL-123 | رجل رجال |

When triliteral nouns have a long vowel in the singular pattern, it may occur in any of the two positions between root letters:

| | gloss | sing. | plural | TM patterns | PRIM codes | Arabic |
|---|---|---|---|---|---|---|
| 11 | friend | SaAoHib | OaSoHaAob | FaaEiL-OaFoEaaL | FvvEvL-OaFoEaaL-123 | صاحب أصحاب |
| 12 | film | fiyolom | OafolaAom | N/A -OaFoEaaL | FvvEvL-OaFoEaaL-123 | فيلم أفلام |
| 13 | book | kitaAob | kutub | FiEaaL-FuEuL | FvEvvL-FuEuL-123 | كتاب كتب |
| 14 | messenger | rasuwol | rusul | FaEuuL-FuEuL | FvEvvL-FuEuL-123 | رسول رسل |
| 15 | road | sabiyol | subul | FaEiiL-FuEuL | FvEvvL-FuEuL-123 | سبيل سبل |

The Arabic word for ‘film’ (12) is a loan world, so the pattern of the singular is anomalous and not listed in TM. The 5 cases are conflated to 2 singular-pattern codes.

In quadriliteral nouns, a long vowel may occur after the third root letter of the singular, or sometimes after the second:

| | gloss | sing. | plural | TM patterns | PRIM codes | Arabic |
|---|---|---|---|---|---|---|
| 16 | statue | timovaAol | tamaAoviyol | FaEoLaaB-FaEaaLiiB | FvEvLvvB-FaEaaLiiB-1234 | تمثال تماثيل |
| 17 | bird | EaSofuwor | EaSaAofiyor | FaEoLuuB-FaEaaLiiB | FvEvLvvB-FaEaaLiiB-1234 | عصفور عصافير |
| 18 | light | qanodiyol | qanaAodiyol | FaEoLiiB-FaEaaLiiB | FvEvLvvB-FaEaaLiiB-1234 | قنديل قناديل |
| 19 | bishop | muToraAon | mataAorinap | FuEoL**aa**B-FaEaaLiBap | FvEvLvvB-FaEaaLiBap-1234 | مطران مطارنة |
| 20 | sample | namuwoJaj | namaAoJij | FaEuuLaB-FaEaaLiB | FvEvvLvB-FaEaaLiB-1234 | نموذج نماذج |

4.1.2. Omission of suffixes

Some singular nouns have a suffix which disappears in the plural. Traditional morphology includes this singular suffix in the singular pattern:

| | gloss | sing. | plural | TM patterns | PRIM codes | Arabic |
|---|---|---|---|---|---|---|
| 21 | knot | Euqod**ap** | Euqad | FuEoL**ap**-FuEaL | FvEvL-FuEaL-123 | عقدة عِقد |
| 22 | bomb | qunobul**ap** | qanaAobil | FuEoLuB**ap**-FaEaaLib | FvEvLvB-FaEaaLiB-1234 | قنبلة قنابل |
| 23 | school | madoras**ap** | madaAoris | maFoEaL**ap**-maFaaEiL | FvEvLvB-FaEaaLiB-1234 | مدرسة مدارس |
| 24 | whore | MaromuwoT**ap** | MaraAomiyoT | FaEoLuuB**ap**-FaEaaLiib | FvEvLvvB-FaEaaLiiB-1234 | شرموطة شراميط |

Such information is unnecessary for producing the broken plural, since the suffix is absent from it. Our model does not specify the suffix in the singular-pattern code, which is generally conflated with a code for nouns without suffix in the singular. This simplification of the BP taxonomy affects many lexical items. The suffix *-ap* is generally the singular suffix for feminine forms (21-24).

The suffix *-iyG* and its feminine counterpart *-iyGap* are typical singular suffixes for human nouns (and adjectives) derived from nouns. Most of such nouns and adjectives pluralize with a sound plural suffix such as *-uwona* or *-aAoT*, but others take a BP:

| | gloss | sing. | plural | TM patterns | PRIM codes | Arabic |
|---|---|---|---|---|---|---|
| 25 | soldier | junod**iyG** | junuwod | FuEoL**iyy**-FuEuuL | FvEvL-FuEuuL-123 | جنديَ جنود |
| 26 | copt | quboT**iyG** | OaqobaAoT | FuEoL**iyy**-OaFoEaaL | FvEvL-OaFoEaaL-123 | قبطيَ أقباط |
| 27 | foreigner | Oajonab**iyG** | OajaAonib | FaEoLaB**iyy**-FaEaaLiB | FvEvLvB-FaEaaLiB-1234 | أجنبيَ أجانب |
| 28 | barbar | barobar**iyG** | baraAobirap | FaEoLaB**iyy**-FaEaaLiBap | FvEvLvB-FaEaaLiBap-1234 | بربريّ برابرة |
| 29 | zionist | Sahoyuwon**iyG** | SahaAoyinap | FaEoLuuB**iyy**-FaEaaLiBap | FvEvLvvB-FaEaaLiBap-1234 | صهيونيَ صهاينة |

The following non-derived nouns illustrate the same situation:

| | gloss | sing. | plural | TM patterns | PRIM codes | Arabic |
|---|---|---|---|---|---|---|
| 30 | rifle | bunoduq**iyGap** | banaAodiq | FuEoLuB**iyyap**-FaEaaLiB | FvEvLvB-FaEaaLiB-1234 | بُنْدُقِيَّة بَنادِق |
| 31 | turtle | suloHaf**aAop** | salaAoHif | FuEoLaB**aap**-FaEaaLiB | FvEvLvB-FaEaaLiB-1234 | سُلَحْفاة سَلاحِف |

**4.2. Broken-plural patterns**

Most BP patterns in the PRIM taxonomy are the same as in traditional morphology. However, a few differences come from our choice to handle patterns and roots at the surface level.

The BP of *miqaSG* ‘scissors’ has two occurrences of the same consonant separated by *i*:

| | gloss | sing. | plural | TM patterns | PRIM codes | Arabic |
|---|---|---|---|---|---|---|
| 32 | scissors | miqaSG | maqaAoSiS | miFaEoL-maFaaEiL | FvEvLvB-FaEaaLiB-1233 | مَقصّ مَقاصص |

Some nouns behave in the same way, except that the two occurrences of the consonant are optionally (33-34) or obligatorily (35-36) replaced by a geminated consonant:

| | gloss | sing. | plural | TM patterns | PRIM codes | Arabic |
|---|---|---|---|---|---|---|
| 33 | porcupine | lutunGap | lataAonin | FuEuLLap-FaEaaLiL | FvEvL-FaEaaLiB-1233 | لتنَّة لتانِن |
| 34 | porcupine | lutunGap | lataAonG | FuEuLLap-FaEaaLiL | FvEvL-FaEaaLiB-123G | لتنَّة لتانّ |
| 35 | mission | muhimGap | mahaAomG | muFoEiLap-maFaaEiL | FvEvLvB-FaEaaLiB-123G | مهمَّة مَهامّ |
| 36 | substance | maAod**G**ap | mawaAodG | FaaEiLap-FawaaEiL | FvvEvL-FaEaaLiB-1w22 | مادَّة مواذّ |

Traditional morphology views these forms as the result of the application of a rule that erases *i* between two occurrences of the same consonant: “The plural *mawaadd* is the form that the plural pattern *fawaaEil* takes in geminate nouns because of the phonological restriction on sequences that include a vowel between identical consonants: **mawaadid* –> *mawaadd*. It is diptote (CaCaaCiC pattern)” (Ryding, 2005:471). In fact, the conditions of application of the rule are also lexical: it does not apply in (e), while it applies optionally in (33-34) and obligatorily in (35-36). Therefore, we account for this morphophonological variation through inflectional classes.

In the BP of (34)-(35), the surface pattern actually handled by the PRIM transducers, *FaEaaLoB*, differs from the traditional deep patterns which contain *i*. In this case, our option for surface patterns tends to increase the number of distinct patterns, and to separate (33)-(35) from (32) in the pattern taxonomy. In order to avoid this effect, we included the deep pattern **label** *FaEaaLiB* in the PRIM inflectional codes. Thus, they sound more familiar to Arabic speakers, because they comply with the deep patterns of traditional morphology taught in school. The forbidden, optional or obligatory geminated consonant is encoded by the respective root codes *1233*, *123G* and *1w22*.

When the BP surface patterns differ from traditional deep patterns, because of morphophonological constraints or variations, the deep pattern **label** is used in the inflectional code, and the surface pattern in the transducer associated to it. Thus, the pattern labels used in inflectional codes are relatively intuitive.

In the following case, we use the same method to conflate BP patterns labels. Some triliteral lemmas have the suffix *-iy* appended to the root in the BP:

| | gloss | sing. | plural | TM patterns | PRIM codes | Arabic |
|---|---|---|---|---|---|---|
| 37 | night | layolap | layaAol**iy** | FaEoLap-FaEaaLiy | FvEvL-FaEaaLiB-123y | ليلة ليالي |

The following nouns are similar, except for a free variation between *-iy* and *-aY*, where *Y* is an allograph of final *A*:

| | gloss | sing. | plural | TM patterns | PRIM codes | Arabic |
|---|---|---|---|---|---|---|
| 38 | desert | SaHoraAoc | SaHaAor**iy** | FaEoLaac-FaEaaLiB | FvEvL-FaEaaLiB-123y | صحراء صحاري |
| 39 | desert | SaHoraAoc | SaHaAor**aY** | FaEoLaac-FaEaaLaY | FvEvL-FaEaaLiB-123Y | صحراء صحارى |
| 40 | complaint | MakowaY | MakaAow**iy** | FaEoLaY-FaEaaLaY | FvEvL-FaEaaLiB-123y | شكوى شكاوي |
| 41 | complaint | MakowaY | MakaAow**aY** | FaEoLaY-FaEaaLaY | FvEvL-FaEaaLiB-123Y | شكوى شكاوى |

The noun *EaJoraAoc* ‘virgin’ has obligatorily *-aY*:

| | gloss | sing. | plural | TM patterns | PRIM codes | Arabic |
|---|---|---|---|---|---|---|
| 42 | virgin | EaJoraAoc | EaJaAor**aY** | FaEoLaac-FaEaaLaY | FvEvL-FaEaaLiB-123Y | عذراء عذارى |

The BP surface pattern actually handled in the implementation details of the PRIM transducers for (39), (41) and (42) is *FaEaaLaB*. However, it is natural to Arabic speakers to consider it as a superficial allomorph of *FaEaaLiB*, which is a regular BP pattern: the fact that the sequence *iY* cannot occur in Arabic explains the surface forms in *aY*. We adopted the pattern label *FaEaaLiB* in the inflectional code, in order to reduce the number of pattern labels and to keep the encoding of these nouns intuitive. The quality of the long vowel in the suffix is encoded in the root code *123Y*.

The same situation occurs in the following examples, with the suffixes *-aAon* in the singular and *-aY* in the BP:

| | gloss | sing. | plural | TM patterns | PRIM codes | Arabic |
|---|---|---|---|---|---|---|
| 43 | drunk | sakor**aAon** | sakaAor**aY** | FaEoLaan-FaEaaLaY | FvEvL-FaEaaLiB-123Y | سكران سكارى |
| 44 | Christian | naSor**aAoniyG** | naSaAor**aY** | FaEoLaaniyy-FaEaaLaY | FvEvL-FaEaaLiB-123Y | نصرانيّ نصارى |

and in the following BPs with the -*aA* ending:

| | | | | | | |
|---|---|---|---|---|---|---|
| 45 | corner | zaAowiyap | zawaAoyaA | FaaEiLap-FaEaayaA | FvvEvL-FaEaaLiB-12yA | زاوية زوايا |
| 46 | mirror | miroCp | maraAoyaA | miFoEaLap-maFaayaA | FvEvL-FaEaaLiB-12yA | مرآة مرايا |
| 47 | intention | niyGap | nawaAoyaA | FiEoLap-FaEaayaA | FvEvL-FaEaaLiB-1wyA | نيّة نوايا |
| 48 | feature | miyozap | mazaAoyaA | FiEoLap-FaEaayaA | FvEvL-FaEaaLiB-13yA | ميزة مزايا |

The sequence *iA* cannot occur in Arabic, which explains the surface forms in *aA*. The quality of the long vowel in the suffix is encoded in the root codes. In example (46), the character *C* ( آ ) is an obligatory substitute for the sequence *OaAo*.

Example (37) poses a segmentation problem. Recall that TM, most analysers and PRIM exclude from the pattern the case and definiteness suffixes. PRIM appends these suffixes to the root/pattern combination during the generation of inflected forms (cf. Section 8.2). In general, these suffixes have little variation depending on lexical entries, and little interaction with the end of the root and pattern. In the case of (37) *layaAoliy* 'nights', the *iy* ending is removed in the indefinite nominative and genitive *layaAolK*. We consider the *iy* ending as a part of the pattern; this ending is removed when the case and definiteness suffixes are appended. Our segmentation is conforted by the fact that in other nouns, *iy* is actually part of the root, as in *qaAoDiy* 'judge' which declines as *qaAoDK* in the indefinite nominative. Our analysis deviates slightly from tradition and simplifies it. According to TM, *iy* is present in underlying forms **layaAoliyN* and **layaAoliyK*, which are both rewritten as the surface form *layaAolK*, and the 'citation form' used to refer to the word is *layaAolK*, a form without *iy*.

### 4.3. Simultaneous conflation of singular and broken-plural patterns

In the framework of traditional morphology, the analysis of broken plurals is systematically consistent with the roots traditionally used for the practical purpose of indexing dictionaries. For instance, the BP of the derived noun *miEowal* 'mattock' is analysed with the root of its derivational base, here *Ewl*. An inflectional phenomenon is thus analysed with a derivational concept. By imposing one of the pieces of the jigsaw (the root), this practice constrains all others, and happens to blur regularities in the system of inflectional patterns.

For the PRIM model, the objective of consistency with derivational analyses is only secondary to the simplicity of the taxonomy. By relaxing this constraint, we can capture more of the regularity of the inflectional system.

#### 4.3.1. Nouns with *m*- prefixes

Many nouns have a *ma*-, *mu*- or *mi*- prefix before a triliteral root. Traditional morphology excludes these prefixes from the root, and consequently includes them in the pattern, on the basis of the derivational history of these words:

| | | | | | | |
|---|---|---|---|---|---|---|
| 49 | mattock | miEowal | maEaAowil | **mi**FoEaL-**ma**FaaEiL | FvEvLvB-FaEaaLiB-1234 | معول معاول |

The prefix is common to the singular and BP of the derived noun. If we analyse the initial *m*- as a part of a quadriliteral root, most of these nouns enter in independently existing inflectional classes. 'Initial *m(i)*-, although originally a prefix, is annexed to the root and treated as a C1 as far as BP formation is concerned' (Kihm, 2006:83). For PRIM, the 9 prefixed nouns below inflect exactly like (A) or (B):

| | | | | | | |
|---|---|---|---|---|---|---|
| **A** | **dagger** | **xanojar** | **xanaAojir** | **FaEoLaB-FaEaaLiB** | **FvEvLvB-FaEaaLiB-1234** | خنجر **خناجر** |
| 50 | theater | masoraH | masaAoriH | **ma**FoEaL-**ma**FaaEiL | FvEvLvB-FaEaaLiB-1234 | مسرح مسارح |
| 51 | house | manozil | manaAozil | **ma**FoEiL-**ma**FaaEiL | FvEvLvB-FaEaaLiB-1234 | منزل منازل |
| 52 | museum | mutoHaf | mataAoHif | **mu**FoEaL-**ma**FaaEiL | FvEvLvB-FaEaaLiB-1234 | متحف متاحف |
| 53 | sieve | munoxul | manaAoxil | **mu**FoEuL-**ma**FaaEiL | FvEvLvB-FaEaaLiB-1234 | منخل مناخل |
| 54 | pulpit | minobar | manaAobir | **mi**FoEaL-**ma**FaaEiL | FvEvLvB-FaEaaLiB-1234 | منبر منابر |
| **B** | **cluster** | **Eunoquwod** | **EanaAoqiyod** | **FuEoLuuB-FaEaaLiiB** | **FvEvLvvB-FaEaaLiiB-1234** | عنقود عناقيد |
| 55 | letter | makotuwob | makaAotiyob | maFoEuuL-maFaaEiiL | FvEvLvvB-FaEaaLiiB-1234 | مكتوب مكاتيب |
| 56 | gutter | mizoraAob | mazaAoriyob | miFoEaaL-maFaaEiiL | FvEvLvvB-FaEaaLiiB-1234 | مزراب مزاريب |
| 57 | poor | misokiyon | masaAokiyon | miFoEiiL-maFaaEiiL | FvEvLvvB-FaEaaLiiB-1234 | مسكين مساكين |
| 58 | napkin | minodiyol | manaAodiyol | miFoEiiL-maFaaEiiL | FvEvLvvB-FaEaaLiiB-1234 | منديل مناديل |

The only reasons to discriminate them are alien to inflectional morphology. In the traditional analysis, both the singular and BP patterns explicitly contain the prefix, which makes them specific to this set of nouns. Even if we strip the prefix off the patterns, we do not always obtain triliteral patterns observable in other BP nouns. Therefore, the traditional analysis increases the number of patterns. By implementing the alternative analysis, PRIM conflates simultaneously the singular pattern and the BP pattern with those of (A) or (B), which simplifies the taxonomy.

The following examples are less regular, but also follow independently observed patterns:

```
59 building    mabonaY    mabaAoniy   maFoEaL-maFaaEiL      FvEvLvB-FaEaaLiB-123y   مبنى مباني
60 school      madorasap  madaAoris   maFoEaLap-maFaaEiL    FvEvLvB-FaEaaLiB-1234   مدرسة مدارسة
61 tragedy     maOosaAop  maCaAosiy   maFoEaLap-maFaaEiL    FvEvLvB-FaEaaLiB-1h3y   مأساة مآسي
62 foreigner   OajonabiyG OajaAonib   FaEoLaBiyy-FaEaaLiB   FvEvLvB-FaEaaLiB-1234   أجنبي أجانب
63 appointment mawoEid    mawaAoEiyod maFoEiL-maFaaEiiL     FvEvLvB-FaEaaLiiB-1234  مَوْعِد مَواعِيد
64 starling    zurozur    zaraAoziyor FuEoLuB-FaEaaLiiB     FvEvLvB-FaEaaLiiB-1234  زرزور زرازير
```

In example (61), the character *C* ( آ ) is an obligatory substitute for the sequence *OaAo*. The morphology of nouns with *ma*-, *mu*- or *mi*- prefix relates them with verb participles. Their derivational patterns are traditionally labelled with semantic features of patient, e.g. *[ktb & maFoEuuL]= makotuwob* مكتوب ‘letter’, derived from the triliteral root *ktb* ‘write’, or of instrument, e.g. *[zrb & miFoEaaL] = mizoraAob* مزراب ‘gutter’, derived from *zrb* ‘flow’. Some of these nouns denote places, e.g. *[nzl & maFoEiL] = manozil* ‘house’ منزل , from *nzl* ‘go down’.

4.3.2. Other cases of diachronically motivated morphological segmentation

In a similar way, some nouns with 4 consonants are traditionally analysed as triliteral, by assigning one of the consonants to the pattern, usually because of a diachronical relation of the noun with a triliteral root, or for some other etymological reason. These nouns can usually be traced back to roots through derivational patterns for participles, deverbal nouns, instrumental nouns… The consonants thus discarded from the root are often *s*, *n*, *t*, *h*, *m*, *w*, *y* or the glottal stop [ʔ], noted by the allographs *c*, *O*, *e*, *W* and *I* (ء, أ, ئـ, ؤ , إ) depending on context. Some of these consonants are more likely to be discarded if they occur in some position in relation to the root. We list below 8 examples of such nouns. If analysed as quadriliteral, all enter in the independently existing inflectional class of *TarobuwoM* ‘tarboosh’ (C), just as if they were synchronically reanalysed as quadriliteral nouns for inflectional purposes:

```
   gloss      singular     plural      TM patterns         PRIM codes                Arabic

C  tarboosh   TarobuwoM TaraAobiyoM FaEoLuuB-FaEaaLiiB FvEvLvvB-FaEaaLiiB-1234   طربوش طرابيش
65 expression taEobiyor taEaAobiyor taFoEiiL-taFaaEiiL FvEvLvvB-FaEaaLiiB-1234   تعبير تعابير
66 week       OusobuwoE OasaAobiyoE OuFoEuuL-OaFaaEiiL FvEvLvvB-FaEaaLiiB-1234   أسبوع أسابيع
67 pumpkin    yaqoTiyon yakaAoTiyon yaFoEiiL-yaFaaEiiL FvEvLvvB-FaEaaLiiB-1234   يقطين يقاطين
68 nostril    xayoMuwom xayaAoMiyom FayoEuuL-FayaaEiiL FvEvLvvB-FaEaaLiiB-1234   خيشوم خياشيم
69 pig        xanoziyor xanaAoziyor FanoEiiL-FanaaEiiL FvEvLvvB-FaEaaLiiB-1234   خنزير خنازير
70 address    EunowaAon EanaAowiyon FuEowaaL-FaEaawiiL FvEvLvvB-FaEaaLiiB-1234   عنوان عناوين
71 coffin     taAobuwot tawaAobiyot FaEoLuut-FaEaaLiit FvEvLvvB-FaEaaLiiB-1234   تابوت توابيت
72 plant      rayoHaAon rayaAoHiyon FaEoLaan-FaEaaLiin FvEvLvvB-FaEaaLiiB-1234   ريحان رياحين
```

(65) is a deverbal noun with the derivational pattern *taFoEiiL* related to the verbal pattern *FaEEaLa*.

(66) *OusobuwoE* أسبوع ‘week’ is related to *saboE* سبع ‘seven’.

In (67) and (68), *y* is considered exterior to the root, probably for some etymological reason.

In (69) *xanoziyor* خنزير ‘pig’, there is no agreement in traditional dictionaries such as Ibn Manzur (1290) and Al-Fairuzabadi (v. 1400): dictionaries consider the *n* in this word as a root consonant or not, because an *n* after the 1st root letter may have a special value.

In (70), *w* after the 2nd root letter may have a special value, and *EunawaAon* ‘address’ may be related to the triliteral root *EnY* عنى ‘signify’.

(71) ends with -*uwot*, a suffix of Aramaic origin, so the final *t* is not considered a root consonant. However, Tarabay (2003) classifies it in both *FaEoLuut-FaEaaLiit* and *FaEoLuuB-FaEaaLiiB*.

In (72), -*aAon* is a suffix, so the final *n* is not considered a root consonant.

The assignment of a consonant to the patterns by traditional morphology makes the patterns of examples (68-70) distant from typical inflectional patterns for nouns, in which phonetic consonants sometimes occur before the 1st root letter, as in *OaFoEaaL* (cf. (11-12), Section 4.1.1), or after the last, as in *FaEoLap* (cf. (36), Section 4.2), but not between root letters, be it in the singular or in the BP. In the PRIM taxonomy, we analyse (65)-(72) as quadriliteral as far as inflection is concerned.

## 5. Root alternations

The root letters of most BPs have the same surface form as those of the singular, as in *Euqodap* 'knot' vs. *Euqad* 'knots'. Other BPs show root alternations, i.e. changes in the surface value of root letters, as in *qabow* 'cave' vs. *Oaqobiyap* 'caves', or in the number of root letters, as in *TaAobiE* 'stamp' طابع vs. *TawaAobiE* 'stamps' طوابع . In the PRIM model, root alternations are represented by a mapping between surface roots from the singular to the BP. This mapping is specified in a straightforward way by **root codes**, a new device.

### 5.1. Bypassing deep roots and rules

In traditional morphology, most root alternations are obtained by applying rules to deep stems. This model has two major drawbacks. First, rules are not very adequate for a phenomenon with such lexical dependency as BP; the few authors that formalized the rules of traditional morphology (Beesley, 1996; Habash, Rambow, 2006; Smrž, 2007) did not publish them in a readable, updatable way. Second, deep roots are not directly observable, which complicates decisions about what their exact value should be. We abandoned this model for root codes, a new device that simplifies the encoding of lexical items, as the following examples show.

#### 5.1.1. Morphophonological alternations of the 2nd root letter

Some nouns with BP are analysed with their 2nd root letter realised as *A* in the singular, and as *w* or *y* in the plural:

| | gloss | sing. | plural | root and patterns (TM) | PRIM codes | in Arabic |
|---|---|---|---|---|---|---|
| 73 | door | baAob | OabowaAob | bwb FaEoL-OaFoEaaL | FvEvL-OaFoEaaL-1w3 | أبواب باب |
| 74 | tooth | naAob | OanoyaAob | nyb FaEoL-OaFoEaaL | FvEvL-OaFoEaaL-1y3 | أنياب ناب |

Traditional morphology describes this with the aid of a deep root, displayed in the examples above just before the TM patterns: *bwb*, *nyb*. In the deep root, the 2nd root letter is the consonant observed in the plural and in derived words. Morphophonological rules change this letter to *A* in the singular, and leave it unchanged in the plural.

In the PRIM model, we specify the presence of *w* or *y* as the 2nd letter of the surface form of the BP root, through the root codes displayed in the examples above at the end of the PRIM codes: *1w3*, *1y3*. The surrounding slots are represented in the root code, as usual, by a digit corresponding to their rank. We stick to directly observable facts. The transducer associated to the inflectional code generates *w* or *y* at the position of the 2nd letter root in the BP. The root code specifies the value of BP root letters when they differ from the corresponding singular root letters. As a simplification, the value of the 2nd letter in the plural is encoded in the root code whenever it is *y*, *w*, a glottal stop [ʔ], or *A*. This is not strictly necessary for the generation of the plural of *suwor* 'wall', which is *OasowaAor*, since root code *123* would yield the same result as *1w3*, but it simplifies the manual encoding of entries.

The following example illustrates the converse situation. The 2nd root letter *y* is replaced by *A* in the plural:

| | gloss | singular | plural | TM root and patterns | PRIM codes | in Arabic |
|---|---|---|---|---|---|---|
| 75 | politician | siyaAosiyG | saAosap | sys FiEaaLiyy-FaAoLap | FvEvvL-FaEolap-1A3 | سياسيَ ساسة |

When the 2nd root letter of a triliteral noun is realised in the singular as [ʔ], the corresponding letter in the plural may be, unpredictably, [ʔ], *y*, *w* or *A*:

| | gloss | singular | plural | TM root and patterns | PRIM codes | in Arabic |
|---|---|---|---|---|---|---|
| 76 | sad | baAoeis | baOasap | bcs FaaEiL-FaEaLap | FvvEvL-FaEaLap-1h3 | بائس بأسة |
| 77 | betrayer | xaAoein | xawanap | xwn FaaEiL-FaEaLap | FvvEvL-FaEaLap-1w3 | خائن خونة |
| 78 | undecided | HaAoeir | Hayarap | Hyr FaaEiL-FaEaLap | FvvEvL-FaEaLap-1y3 | حائر حيرة |
| 79 | seller | baAoeiE | baAoEap | byE FaaEiL-FaEoLap | FvvEvL-FaEoLap-1A3 | بائع باعة |

The letters *c* and *O* note allographs of the glottal stop [ʔ]. Traditional morphology postulates deep roots. In (79), the underlying *y* of the deep root occurs neither in the singular nor in the plural; rules change it to *e* in the singular and to *A* in the BP.

We encode the 2nd root letter of the plural in the root code: *1h3*, *1w3*, *1y3*, *1A3*. In root codes, the symbol *h* stands for [ʔ]. There are much less distinct root codes in the PRIM model than roots in TM: all the deep roots of triliteral nouns with alteration of the 2nd root letter conflate to the 4 code roots cited above.

5.1.2. Morphophonological alternations of the 3rd root letter

The situation is the same for nouns which alter their 3rd root letter. In the BP, this letter is realised as *y* or *c*, or as the long vowel [a:], noted *A* or *Y*:

| | gloss | sing. | plural | TM root and patterns | PRIM codes | in Arabic |
|---|---|---|---|---|---|---|
| 80 | organ | EuDow | OaEoDaAoc | ED- FuEow-OaFoEaaL | FvEvL-OaFoEaaL-12h | عُضْو أَعْضاء |
| 81 | cloth | zayG | OazoyaAoc | zy- FaEE-OaFoEaaL | FvEvL-OaFoEaaL-12h | زِيّ أَزْياء |
| 82 | climate | jawG | OajowaAoc | jw- FaEE-OaFoEaaL | FvEvL-OaFoEaaL-12h | جوّ أجواء |
| 83 | enemy | EaduwG | OaEodaAoc | Ed- FaEuuw-OaFoEaaL | FvEvvL-OaFoEaaL-12h | عدوّ أعداء |
| 84 | cave | qabow | Oaqobiyap | qb- FaEow-OaFoEiLap | FvEvL-OaFoEiLap-12y | قبو أقبية |
| 85 | pot | wiEaAoc | OawoEiyap | wE- FiEaac-OaFoEiLap | FvEvvL-OaFoEiLap-12y | وعاء أوعية |
| 86 | boy | fataY | futoyaAon | ft- FaEaY-FuEoLaan | FvEvL-FuEoLaan-12y | فتى فتيان |
| 87 | boy | fataY | fitoyap | ft- FaEaY-FiEoLap | FvEvL-FiEoLap-12y | فتى فتية |
| 88 | judge | qaAoDiy | quDaAop | qD- FaaEiy-FuEaap | FvvEvL-FuEoLap-12A | قَاضِي قُضَاة |
| 89 | jewel | Hiloyap | HilaY | Hl- FiEoyap-FaEaY | FvEvL-FiEaL-12Y | حلية حلًى |
| 90 | step | xuTowap | xuTaY | xT- FuEowap-FuEaY | FvEvL-FuEaL-12Y | خطوة خطى |

Since scholars may disagree on the value of the 3rd letter of the traditional deep root, we omit it above. In the PRIM model, the surface value of the 3rd root letter in the plural is encoded in the root code whenever it is *y*, [ʔ], *A* or *Y*:

| | gloss | sing. | plural | TM root and patterns | PRIM codes | in Arabic |
|---|---|---|---|---|---|---|
| 91 | valley | waAodiy | Oawodiyap | wd- FaaEiL-OaFoEiLap | FvvEvL-OaFoEiLap-12y | وادي أودية |
| 92 | pastor | raAoEiy | ruEoyaAon | rE- FaaEiL-FuEoLaan | FvvEvL-FuEoLaan-12y | راعي رعيان |

5.1.3. Orthographic alternations of glottal stop in roots

Roots with the glottal stop [ʔ] undergo purely orthographic alternations. The glottal stop [ʔ] has 6 allographs in the Arabic alphabet: *c*, *e*, *W*, *O*, *I* and *C* (ء, أ, ئـ, ؤ , إ, آ) . In general, the choice of the allograph depends on orthographic context, and in particular on the preceding and following vowels.[17] For example, an initial [ʔ] is written *O* ( أ ) when it is followed by *a* or *u*, and *I* ( إ ) when followed by *i*. The character *C* ( آ ) is an obligatory substitute for the sequences *OaAo* and *OaOo*. The allographs can be different between the singular and the plural, because they are inserted in different patterns:

| | gloss | singular | plural | TM root and patterns | PRIM codes | in Arabic |
|---|---|---|---|---|---|---|
| 93 | kettle | Iiboriyoq | **O**abaAoriyoq | IiFoEiiL-OaFaaEiiL | FvEvLvvB-FaEaaLiiB-h234 | إبريق أباريق |
| 94 | African | IiforiyoqiyG | **O**afaAoriqap | IiFoEiiLiyG-OaFaaEiLap | FvEvLvvB-FaEaaLiBap-h234 | إفريقي أفارقة |

Because of these spelling changes, we systematically register in root codes the presence of [ʔ]. In root codes, the symbol *h* stands for [ʔ]. Then, the plural pattern is sufficient to determine the allograph in the BP:

[17] In some configurations, no standard is actually applied to determine the allograph, and practice depends on regions and authors. In Arabic dialects, initial [ʔ] admits phonetic variants, and some of them may have an influence on spelling in Modern Standard Arabic.

| | | | | | |
|---|---|---|---|---|---|
| 95 trouble | **ma**Oozaq | maCziq | **ma**FoEaL-**ma**FaaEiL | FvEvLvB-FaEaaLiB-1h34 | مأزق مآزق |
| 96 twin | ta**w**o**O**am | tawaAo**e**im | Fa**w**oEaL-Fa**w**aaEiL | FvEvLvB-FaEaaLiB-12h4 | توءم توائم |
| 97 congrat. | tahonieap | tahaAonie | taFoEiLap-taFaaEiL | FvEvLvB-FaEaaLiB-123h | تهنئة تهانئ |
| 98 principle | **ma**bodaO | mabaAodie | **ma**FoEaL-**ma**FaaEiL | FvEvLvB-FaEaaLiB-123h | مبدأ مبادئ |
| 99 pearl | luWoluW | laClie | FuEoFuE-FaEaaFiE | FvEvLvB-FaEaaLiB-1**h**3**h** | لؤلؤ لآلئ |

The correct allograph of [ʔ] is inserted by the transducer associated to the inflectional code. It is not necessary to specify it in the root code, since it depends on the context, which is encoded in the BP pattern.[18]

Even when the allograph is the same in the singular and in the plural, we encode the presence of the glottal stop in the root code (100, 101). This is not strictly necessary for the generation of the plural, since in such case root code *1234* would yield the same result as *h234*, but it simplifies the manual encoding of entries:

| | | | | | |
|---|---|---|---|---|---|
| 100 warehouse | OanobaAor | OanaAobir | OaFoEaaL-OaFaaEiL | FvEvLvvB-FaEaaLiB-h234 | أنبار أنابير |
| 101 teacher | OusotaAoJ | OasaAotiJap | OuFoEaaL-OaFaaEiLap | FvEvLvvB-FaEaaLiBap-h234 | أستاذ أساتذة |

The allography of [ʔ] poses problems in stem-final position. The allograph may depend on graphically agglutinated pronouns:

| | | |
|---|---|---|
| ruWasaAoci | 'presidents' | رؤساء |
| ruWasaAoeihaA | 'its presidents' | رؤسائها |

In these examples, the final *i* is an inflectional suffix and *-haA* is a clitic pronoun in the genitive. This problem is dealt with in Section 8.

Nouns with initial [ʔ] and BP pattern *OaFoEaaL* pose another problem of allography. In the plural, the combination of the root with the pattern produces an underlying form that begins with the sequence *OaOo*. Due to morphophonological rules, this initial sequence is not pronounced [ʔaʔ] but [ʔa:], and the surface form is not scripted *OaOo* or *OaAo*, but *C* آ :

| | | | | | |
|---|---|---|---|---|---|
| 102 horizon | Oufuq | CfaAoq | FuEuL-OaFoEaaL | FvEvL-OaFoEaaL-h23 | أفق آفاق |

The PRIM transducers actually produce *C*, but we named the root code *h23* and not *A23*, to remind the underlying [ʔ]: since words in Arabic never begin with a long vowel, it is not natural to Arabic speakers to consider that a root begins with *A*.

5.1.4. Biliteral nouns

There are less than 20 biliteral nouns in Arabic. When they admit a BP, it is always triliteral, often with the addition of a final consonant, generally *c*:

| gloss | sing. | plural | TM root and patterns | PRIM codes | in Arabic |
|---|---|---|---|---|---|
| 103 blood | dam | dimaAo**c** | dmc FaE-FiEaaL | FvE-FiEaaL-12h | دم دماء |
| 104 father | Oab | CbaAo**c** | Obw FaE-OaFoEaaL | FvE-OaFoEaaL-h2h | أب آباء |
| 105 brother | Oax | Iixo**w**ap | Oxw FaE-FiEoLap | FvE-FiEoLap-h2w | أخ إخوة |

Traditional morphology generally describes such nouns with a triliteral deep root in which the 3rd root letter is not realised in the singular. Some scholars disagree on this notion of false biliteral, and analyse these roots as underlyingly biliteral. The PRIM taxonomy uses a biliteral singular-pattern code.

A small series of nouns begin with *Ii* in the singular,[19] and have two other consonants; this initial part is pronounced only if the word is preceded by a pause:

| | | | | | |
|---|---|---|---|---|---|
| 106 son | Iibon | OabonaAoc | bnc FoE-OaFoEaaL | FvEvL-OaFoEaaL-23h | إبن أبناء |
| 107 name | Iisom | OasomaAoc | smc FoE-OaFoEaaL | FvEvL-OaFoEaaL-23h | إسم أسماء |

According to traditional morphology, this initial letter does not count as a root letter, so these nouns are biliteral. We encode them as triliteral.

[18] (97) admits an alternative plural, *tahaAoniy*, which is assigned to another lexical entry (cf. Section 3.2).
[19] Recall that *I* is an allograph of [ʔ].

### 5.2. Shifting information from broken-plural patterns to root codes

In some cases, traditional morphology accounts for consonant insertions through special BP patterns such as *FawaaEiL*, *FaEaaeiL*, *FaEaayiL* (فواعل, فعائل, فعايل). By encoding such insertions in root codes, we reduce the number of BP patterns.

<u>5.2.1. Triliteral lemmas with insertion of *y*, *w* or [ʔ]</u>

The following nouns have 3 phonetic consonants in the singular, excluding suffixes, and 4 in the BP:

| | gloss | singular | plural | TM patterns | PRIM codes | in Arabic |
|---|---|---|---|---|---|---|
| 108 | stamp | TaAobiE | TawaAobiE | FaaEiL-FawaaEiL | FvvEvL-FaEaaLiB-1w23 | طابع طوابع |
| 109 | order | Oamor | OawaAomir | FaEoL-FawaaEiL | FvEvL-FaEaaLiB-1w23 | أمر أوامر |
| 110 | brothel | maAoxuwor | mawaAoxir | FaaEuuL-FawaaEiL | FvvEvvL-FaEaaLiB-1w23 | ماخُور مَواخِر |
| 111 | last | Cxir | OawaAoxir | FaaEiL-FawaaEiL | FvvEvL-FaEaaLiB-hw23 | آخر أواخر |
| 112 | revenue | EaAoeid | EawaAoeid | FaaEiL-FawaaEiL | FvvEvL-FaEaaLiB-1wh3 | عائد عوائد |
| 113 | darling | Habiyob | HabaAoyib | FaEiiL-FaEaayiL | FvEvvL-FaEaaLiB-12y3 | حبيب حبايب |
| 114 | old | Eajuwoz | EajaAoeiz | FaEuuL-FaEaaeiL | FvEvvL-FaEaaLiB-12h3 | عَجُوز عَجَائِز |
| 115 | first | OawGal | OawaAoeil | FaEEaL-FaEaaeiL | FvEEvL-FaEaaLiB-12h3 | أوّل أوائل |
| 116 | angel | malaAok | malaAoeikap | FaEaaL-FaEaaeiLap | FvEvvL-FaEaaLiBap-12h3 | ملاك ملائكة |

Traditional morphology postulates that the deep root is the same for all the forms of a lexical entry. In consequence, the BP of these nouns has to be analysed with triliteral roots; the additional consonant can only be assigned to the pattern. This generates several additional BP patterns which specify the position and value of the additional consonant, as *FawaaEiL*. The fact that the additional consonant occurs between the slots for root letters in these patterns makes them distant from other inflectional patterns for nouns, as *FaEaaLiB*. Recall that in typical inflectional patterns for nouns, be it in the singular or in the BP, phonetic consonants sometimes occur before the 1st slot, as in *OaFoEaaL*, or after the last, as in *FaEoLap*, but not between slots (Section 4.3.2).

In contrast, if we analyse the nine BPs above (108-116) with quadriliteral roots, all their patterns conflate with *FaEaaLiB* and *FaEaaLiBap*, which are independently needed for other BPs. We adopted this solution for the PRIM taxonomy. We use the root code to specify the insertion of the additional consonant in the plural root. This analysis simplifies the BP pattern taxonomy by merging classes. It changes the BP patterns, but it remains straightforward to Arabic speakers, since it reuses familiar BP patterns.

In these nouns, the position of the additional consonant of the BP is often occupied by a long vowel in the singular. For a couple of them, an alternative analysis is possible, in which the singular has a quadriliteral root, and one of the root letters codes the long vowel of the singular, as in (117a):

| | gloss | singular | plural | TM root and patterns | PRIM codes | in Arabic |
|---|---|---|---|---|---|---|
| 117 | missile | S**aAo**ruwox | SawaAoriyox | Srx FaaEuuL-FawaaEiiL | FvvEvvL-FaEaaLiiB-1w23 | صاروخ صواريخ |
| 117a | | | | Swrx FaEoLuuB-FaEaaLiiB | | |
| 118 | wheel | d**uwo**laAob | dawaAoliyob | dlb FuuEaaL-FawaaEiiL | FvvEvvL-FaEaaLiiB-1w23 | دولاب دواليب |
| 118a | | | | dwlb FuEoLaaB-FaEaaLiiB | | |

The two alternative analyses (117) and (117a) do not correspond to distinct interpretations of the form: they are two formal accounts for a single linguistic object. This situation requires a choice, so that the morphological analysis reports a single analysis. The solution of (117a) has the advantage of being closer to the encoding of lemmas with 2 phonetic consonants, such as *baAob* 'door' (Section 5.1.1). However, we opted for the solution of (117) which is consistent with (108)-(116). The availability of several solutions to describe the same phenomenon is a flaw in a descriptive model. In order to reduce this indeterminacy in the encoding of entries, we adopted the following rule:

> *For nouns with at least 3 phonetic consonants in the singular stem, long vowels occurring between the first 3 consonants are assigned to the pattern.*

For example, as *SaAoruwox* 'missile' has 3 phonetic consonants *S*, *r* and *x*, the long vowel *aA* is assigned to the pattern, which is specified by picking the singular-pattern code *FvvEvvL*. This rule leads to familiar patterns: for example, *FaEaaLiiB*, in (117) and (118), is independently needed for other nouns. The rule does not apply to *baAob* ‘door’ since this noun has only 2 phonetic consonants. In this type of nouns, the long vowel between the two consonants is unanimously analysed as a root letter.

Traditional morphology has still another analysis for similar nouns, adopting the root of their derivational base:

| | gloss | singular | plural | TMroot | patterns | PRIM codes | in Arabic |
|---|---|---|---|---|---|---|---|
| 119 | port | miyonaAoc | ma**w**aAonie | ? | miFoEaaL-maFaaEiL | FvvEvvL-FaEaaLiB-1w2h | ميناء موانئ |
| 120 | scale | miyozaAon | ma**w**aAoziyon | wzn | miFoEaaL-maFaaEiiL | FvvEvvL-FaEaaLiiB-1w23 | ميزان موازين |
| 121 | cave | magaAorap | magaAo**w**ir | gwr | maFoEiLap-maFaaEiL | FvEvvL-FaEaaLiB-12w3 | مَغارة مَغاوِر |
| 122 | defect | maEaAobap | maEaAo**y**ib | Eyb | maFoEaLap-maFaaEiL | FvEvvL-FaEaaLiB-12y3 | مَعابة مَعايِب |

We opted for the solution of (108-116), for the same reasons as in Section 4.3.1. [20]

The noun *EaAodap* ‘habit’ shows, in addition to the insertion of *w* before the 2nd root letter, the substitution of *e* for *A* as 2nd root letter:

| | | | | | |
|---|---|---|---|---|---|
| 123 | habit | E**aAo**dap | Ea**w**aAo**e**id | FaEoLap-FawaaeiL | FvEvL-FaEaaLiB-1wh3 | عادة عوائد |

We have analysed all the nouns in this section with a triliteral root in the singular, and a quadriliteral root in the plural. In the following sections, we survey other examples of this configuration, where the additional root consonant is obtained by reduplicating one of those of the singular, or by inserting a prefix or a suffix. Then, we discuss the case of nouns with 5 consonants in the singular, and 4 in the BP, obtained by removing one of the 5 consonants.

Most quadriliteral BPs show no root alterations as compared to the singular (cf. (16-20), Section 4.1.1). They have one of the three following patterns: *FaEaaLiB*, *FaEaaLiBap* and *FaEaaLiiB*.

### 5.2.2. Triliteral lemmas with geminated consonant and quadriliteral BP

A number of lemmas with a geminated consonant have a quadriliteral BP. In general, the geminated consonant appears in the plural as two simple occurrences, with a long vowel between them:

| | | | | | | |
|---|---|---|---|---|---|---|
| 124 | ladder | su**lG**am | sa**l**aAo**l**im | FuEEaL-FaEaaEiL | FvEEvL-FaEaaLiB-1223 | سُلَّم سَلالِم |
| 125 | pillow | Ta**rG**aAoHap | Ta**r**aAo**r**iyoH | FaEEaaLap-FaEaaEiiL | FvEEvvL-FaEaaLiiB-1223 | طَرّاحَة طراريح |
| 126 | mighty | ja**bG**aAor | ja**b**aAo**b**irap | FaEEaaL-FaEaaEiLap | FvEEvvL-FaEaaLiBap-1223 | جبّار جبابرة |
| 127 | dragon | tinGiyon | tanaAoniyon | FiEEiiL-FaEaaEiiL | FvEEvvL-FaEaaLiiB-1223 | تنّين تنانين |
| 128 | ox | fidGaAon | fadaAodiyon | FiEEaaL-FaEaaEiiL | FvEEvvL-FaEaaLiiB-1223 | فدّان فدادين |
| 129 | needle | dabGuwos | dabaAobiyos | FaEEuuL-FaEaaEiiL | FvEEvvL-FaEaaLiiB-1223 | دبّوس دبابيس |

The geminated consonant of the singular is analysed as a single letter of a triliteral root, and the gemination is assigned to the singular pattern (cf. Section 3.3). The root code *1223* specifies the repetition of the 2nd root letter. In *OawGal* ‘first’, the geminated consonant of the singular is realised as a simple consonant in the plural, but an additional *e* (ئ) is inserted:

| | | | | | | |
|---|---|---|---|---|---|---|
| 130 | first | Oa**wG**al | Oa**w**aAo**e**il | FaEEaL-FaEaaeiL | FvEEvL-FaEaaLiB-12h3 | أوّل أوائل |

In *MidGap* ‘trouble’, the geminated consonant corresponds to two letters of a triliteral root, and an additional *e* is inserted between them:

| | | | | | | |
|---|---|---|---|---|---|---|
| 131 | trouble | Mi**dGap** | Ma**d**aAo**e**i**d** | FiEoLap-FaEaaeiL | FvEvL-FaEaaLiB-12h2 | شدّة شدائد |

Some triliteral nouns have a quadriliteral BP with a reduplication of the 2nd root letter and a long vowel between the two occurrences:

| | | | | | | |
|---|---|---|---|---|---|---|
| 132 | dinar | diyo**n**aAor | da**naAon**iyor | FiiEaaL-FaEaaEiiL | FvvEvvL-FaEaaLiiB-1223 | دينار دنانير |
| 133 | lighthouse | fa**n**aAor | fa**naAon**iyor | FaEaaL-FaEaaEiiL | FvEvvL-FaEaaLiiB-1223 | فَنار فَنانِير |
| 134 | mortar | haAo**w**un | ha**waAow**iyon | FaaEuL-FaEaaEiiL | FvvEvL-FaEaaLiiB-1223 | هاوُن هَواوِين |

These nouns seem to have atypical origins, since they are not related to attested verbal forms.

### 5.2.3. Triliteral lemmas with BP in -*iy* or -*aY*

Some triliteral lemmas have a quadriliteral BP with -*iy* or -*aY* appended to the root (cf. (37)-(42), Section 4.2):

[20] (119) admits an alternative plural, *mawaAoniy*, which is assigned to another lexical entry (cf. Section 3.2).

| | | | | | | |
|---|---|---|---|---|---|---|
| 135 | bottle | qanGiyon**ap** | qanaAon**iy** | FaEEiiLap-FaEaaLiy | FvEEvvL-FaEaaLiB-123y | قنينة قناني |
| 136 | land | OaroD | OaraAoD**iy** | FaEoL-FaEaaLiy | FvEvL-FaEaaLiB-h23y | أرض أراضي |
| 137 | night | layolap | layaAoliy | FaEoLap-FaEaaLiy | FvEvL-FaEaaLiB-123y | ليلة ليالي |
| 138 | snake | OafoE**aY** | OafaAoE**iy** | FaEoLaY-OaFaaEiy | FvEvL-FaEaaLiB-h23y | أفعى أفاعي |
| 139 | virgin | EaJor**aAoc** | EaJaAor**aY** | FaEoLaac-FaEaaLaY | FvEvL-FaEaaLiB-123Y | عذراء عذارى |

In most of these examples, the singular has a suffix such as *-ap* or *-aY*, which suggests that the ending *-iy* is also a suffix. However, by analysing these endings as part of the stem, we homogenize the nouns with other quadriliteral BPs with pattern *FaEaaLiB*.

In the following examples, *y* is the 3rd consonant of the singular root, and a *w* is inserted before the 2nd consonant, as in (108)-(112), Section 5.2.1:

| | | | | | | |
|---|---|---|---|---|---|---|
| 140 | suburb | DaAoHiy**ap** | Da**w**aAoHiy | FaaEiLap-FawaaEiL | FvvEvL-FaEaaLiB-1w2y | ضاحية ضواحي |
| 141 | whore | EaAoriy**ap** | Ea**w**aAoriy | FaaEiLap-FawaaEiL | FvvEvL-FaEaaLiB-1w2y | عارية عواري |

5.2.4. Triliteral lemmas with BP in *Oa-* or *ma-*

Some triliteral nouns have a BP with an initial *Oa-*, often in concurrence with another plural.[21] We encode the BP in *Oa-* as quadriliteral if it matches one of the three independently known quadriliteral BP patterns (143), and as triliteral otherwise (142):

| | gloss | singular | plural | PRIM codes | in Arabic |
|---|---|---|---|---|---|
| 142 | place | makaAon | **Oa**mokinap | FvEvvL-OaFoEiLap-123 | مَكان أمكنة |
| 143 | place | makaAon | **Oa**maAokin | FvEvvL-FaEaaLiB-h123 | مَكان أماكِن |

In TM, the BP in (143) is marked as 'plural of plural' and obtained by re-pluralizing the BP in (142):

| | gloss | singular | plural | pl. of pl. | TM patterns | in Arabic |
|---|---|---|---|---|---|---|
| 144 | place | makaAon | **Oa**mokinap | **Oa**maAokin | FaEaaL-OaFoEiLap-OaFaaEiL | مَكان أمكنة أماكِن |

Recall that we do not formalize the 'plural of plural' mark in our model (cf. Section 3). Here is a similar example, but both BPs have quadriliteral patterns:

| | gloss | singular | plural | PRIM codes | in Arabic |
|---|---|---|---|---|---|
| 145 | pregnant | HabolaY | HabaAolaY | FvEvL-FaEaaLiB-123Y | حبلَى حبالى |
| 146 | pregnant | HabolaY | **Oa**HaAobiyol | FvEvL-FaEaaLiiB-h123 | حبلَى أحابيل |

| | gloss | sing. | plural | pl. of pl. | TM patterns | in Arabic |
|---|---|---|---|---|---|---|
| 147 | pregnant | HabolaY | HabaAolaY | **Oa**HaAobiyol | FaEoLaY-FaEaaLaY-OaFaaEiiL | حبلَى حبالى أحابيل |

The noun *Hadiyov* 'talk' has only one BP in *Oa-*:

| | gloss | singular | plural | TM patterns | PRIM codes | in Arabic |
|---|---|---|---|---|---|---|
| 148 | talk | Hadiyov | **Oa**HaAdiyov | FaEiiL-OaFaaEiiL | FvEvvL-FaEaaLiiB-h123 | حديث أحادِيث |

Finally, some triliteral nouns have a quadriliteral BP with an initial *ma-*:

| | | | | | | |
|---|---|---|---|---|---|---|
| 149 | feeling | MuEuwor | maMaAoEir | FuEuuL-maFaaEiL | FvEvvL-FaEaaLiB-m123 | شعور مشاعر |
| 150 | danger | xaTar | maxaAoTir | FaEaL-maFaaEiL | FvEvL-FaEaaLiB-m123 | خطر مخاطر |
| 151 | drawback | sayGicap | masaAowie | FaEEiLap-maFaaEiL | FvEEvL-FaEaaLiB-m1wh | سَيِّئة مَساوِئ |

Dictionaries describe this type of plural, but grammarians have paid little attention to them. Tarabay (2003) does not mention them. These nouns usually denote abstract entities and are derived from verbs or adjectives. The *ma-* insertion can be compared with *Oa-* and with derivational prefixes in *m-* occurring in past participles and deverbal nouns. Diachronically, the singular and the plural of such pairs may have come from distinct lexical items. However, synchronically, their association within a single item is confirmed by comparing sentences such as:

[21] As a rule, we generate at most one plural of a given lexical entry. When several plurals are observed, they are assigned to distinct entries, no matter whether they are equivalent or not (cf. Section 3.2).

جلس الشيخ في قاعة الإجتماعات يراجع حساباته الانتخابية
*jalasa <u>Al-Mayoxu</u> fiy qaAEapi Al-IijtimaAEaAti yuraAjiEu HisaAbaAti-hi Al-IntixaAbiyap*
sat the-sheikh in the-room-meeting review calculation-his electoral
"The sheikh sat in the meeting room reviewing his electoral calculation"

جلست المشايخ في قاعة الإجتماعات تراجع حساباتها الانتخابية
*jalasat <u>Al-maMaAyixu</u> fiy qaAEapi Al-IijtimaAEaAti turaAjiEu HisaAbaAti-hA Al-IntixaAbiyap*
sat the-sheikhs in the-room-meeting review calculation-her electoral
"The sheikhs sat in the meeting room reviewing their electoral calculations"

The only semantic difference between these two sentences is about the number of the subject. Such differential semantic evaluation (Gross, 1975) is a particularly reliable and reproducible type of introspective evidence about semantic facts.

<u>5.2.5. Lemmas with 5 or 6 consonants</u>

From a 5-consonant singular, the formation of a quadriliteral BP requires the omission of one of the 5 consonants. The first consonant is never omitted. The consonants *y*, *w* or an *n* are often omitted:

152 philosopher
fa**y**olasuwof falaAosifap Fa**y**oEaLuuB-FaEaaLiBap FvEvLvBvvD-FaEaaLiBap-1345 فيلسوف فلاسفة
153 program
baro**n**aAomaj baraAomij FaEo**n**aaLaB-FaEaaLiB FvEvLvvBvD-FaEaaLiB-1245 برنامج برامج
154 elephant (female)
EaqaroTal EaqaAoril FaEaLoBaD-FaEaaLiD FvEvLvBvD-FaEaaLiB-1235 عقرطل عقارل
155 cylinder
OusoTu**w**aAonap OasaAoTiyon FuEoLu**w**aaBap-FaEaaLiiB FvEvLvBvvD-FaEaaLiiB-h235 أسطوانة اساطين

Note that in the singular, for TM, the consonant omitted in the BP is assigned to the pattern in (152, 153, 154), but to the root in (155).

The 5th consonant is often omitted:

156 quince
safaroja**l** safaAorij FaEaLoBaD-FaEaaLiB FvEvLvBvD-FaEaaLiB-1234 سفرجل سفارج
157 octopus
OaxoTabuwo**T** OaxaAoTib FaEoLaBuuD-FaEaaLiB FvEvLvBvvD-FaEaaLiB-h234 أخطبوط أخاطب

Here is a similar example with 6 consonants:

158 emperor Ii**m**oba**r**aAoTuwor OabaAoTirap FvEvLvBvvDvvJ-FaEaaLiBap-h356 إمبراطور أباطرة

A few 5-consonant nouns deviate from the standard quadriliteral BP patterns in that all 5 root consonants are retained in the BP, with the 3rd and 4th ones jointly in the 3rd slot of the BP pattern:

159 crab silo**T**o**E**aAon salaAo**ToE**iyon FvEvLvBvvD-FaEaa**L**iiB-12345 سلطعان سلاطعين
160 pot miro**T**o**b**aAon maraAo**Tob**iyon FvEvLvBvvD-FaEaa**L**iiB-12345 مرطبان مراطبين
161 thimble kiMo**t**o**b**aAon kaMaAo**tob**iyon FvEvLvBvvD-FaEaa**L**iiB-12345 كشتبان كشاتبين

The surface pattern actually handled by the PRIM transducers of these BPs is *FoEaaLoBiiD*. However, we analyse this pattern as a variant of quadriliteral *FaEaaLiiB*, and we use the label of this pattern in the inflectional codes. These nouns deviate from general rules in several ways. First, all other BP roots have at most 4 consonants. Second, these BPs are pronounced in three syllables as Cv-<u>CvvC</u>-CvvC with unusual CvvC second syllables: [sala:tˤʕi:n mara:tˤbi:n kaʃa:tbi:n ʔatˤa:rmi:zˤ], as if the attraction to a quadriliteral BP pattern were stronger than phonotactic constraints. We are not aware of any prior mention of these exceptional nouns in literature about Arabic.

Unlike standard Arabic, we report, in the Lebanese dialect, the existence of initial consonant clusters for examples (159-161) as *solaAoToEiyon*, pronounced in two syllables as CCvvC-CvvC [sla:tˤʕi:n mra:tˤbi:n kʃa:tbi:n]. (163) is a similar example with an initial consonant cluster, but in a triliteral BP pattern; (162) is the BP of this word in standard Arabic. A probable template for (163) in standard modern Arabic is the inflectional class of (164), with a standard BP pattern *FiEaL*:

```
162 strip      MariyoTap-MaraAoeiT      FvEvvL-FaEaaLiB-12e4      شريطة شرائط
163 strip      MoriyoTap-MoriyaT        F1F2vEvL-F1F2iEaL-1y3     شريطة شريط
164 uprising   fitonap-fitan            FvEvL-FiEaL-123           فتنة فتن
```

Two other plurals of the same noun are observed in the Lebanese dialect: a suffixal plural *MoriyoT-aAot* شريطات and a variant of (162), *MaraAoyiT*.

## 6. Quantitative data about the taxonomy

Our BP lexicon is composed of 3 198 noun entries, among which 1 662 admit a triliteral BP, and 1 536 a quadriliteral BP. We have 985 BPs with the *FaEaaLiB* pattern. Table 1 shows how entries with this BP pattern are distributed according to the singular-pattern taxonomy.

| Singular-Pattern Code | | Example | | | Entries | In Arabic script |
|---|---|---|---|---|---|---|
| | | Gloss | Plural | Singular | | |
| FvEvLvB | FvEvLvB | dirham | daraAhim | diroham | 556 | درهم دراهم |
| | FvEvLvB-ap | tornado | zawaABiE | zawobaEap | | زَوْبَعَة زَوأبَع |
| | FvEvLvB-iyy | foreigner | OajaAnib | OajonabiyG | | أجنبي أجانب |
| | FvEvLvB-iyyap | rifle | banaAdiq | bunduqiyGap | | بُنْدُقِيَّة بَنادِق |
| | FvEvLvB-p | turtle | salaAHif | suloHaFaAp | | سُلَحْفاة سَلاحِف |
| FvEvvLvB | | sample | namaAzij | namuwozaj | 1 | نموذج نماذج |
| FvEvLvvB | | bat | waTaAwiT | wuTowaAT | 19 | وَطْواط وَطاوِط |
| FvEvLLvB | | buildings | majaAmiE | mujamGaE | 4 | مجمّع مجامع |
| FvvEvL | | stamp | tawaAbiE | TaAobiE | 165 | طابع طوابع |
| FvEEvvL | | bottle | **qanaAniy** | qanGiynap | 1 | قنّينة قناني |
| FvvEvvL | | port | **mawaAnie** | miyonaAoc | 6 | ميناء موانئ |
| FvEvvL | | cave | magaAwir | magaAorap | 197 | مَغارة مَغاوِر |
| FvEEvL | | ladder | salaAlim | sulGam | 5 | سُلَّم سَلالِم |
| FvEvL | | order | OawaAmir | Oamor | 25 | أمر أوامر |
| FvEvLvBvD | | quince | safaArij | safarojal | 4 | سفرجل سفارج |
| FvEvLvvBvD | | program | baraAmij | baronaAomaj | 1 | برنامج برامج |
| FvEvLvBvvD | | octopus | OaxaATib | OaxoTabuwoT | 1 | أخطبوط أخاطب |
| | | **TOTAL** | | | **985** | |

Table 1. Distribution of lexical items with the *FaEaaLiB* BP pattern according to the singular-pattern taxonomy.

The 3 198 entries with BP are inflected by means of finite-state transducers in number, definiteness and case (3×3×3). An entry which does not inflect in gender produces 27 surface forms. An entry which inflects also in gender produces 2×3×3×2 forms for the singular and the dual, which inflect in gender, and 1×3×3×1 for the BP, which does not inflect in gender (cf. Section 7); this totals to 45. The size of the full-form dictionary is 97 002 surface forms. It occupies 4.9 Megabytes in Unicode little Endian in plain text. It is compressed and minimized into 430 Kilobytes, and loaded to memory for fast retrieval. The generation, compression and minimization of the full-form lexicon lasts a few seconds on a Windows laptop.

The number of inflectional graphs is 300 : 25 BP patterns, 75 singular pattern/BP pattern pairs, 160 singular pattern/BP patterns/root code triples, and 300 when we take into account the generation of gender and inflectional suffixes in the singular. In addition, the main graphs invoke approximately 20 sub-graphs.

This number of inflectional graphs (300) is to be compared with the nearly 390 inflectional graphs for nouns for Brazilian Portuguese constructed also for Unitex (Muniz *et al.*, 2005) which deals with gender, number and degree (base, diminutive and augmentative), as in *casa(s)* 'house(s)', *casinha(s)* 'small house(s)', *casarão/casarões* 'large house(s)'. Another 245 inflectional graphs for adjectives deal with gender, number and degree: *lindo(s)/linda(s)* 'beautiful' (base), *lindinho(s)/lindinha(s)* (diminutive), *lindão/lindões/lindona(s)* (augmentative) and *lindíssimo(s)/lindíssima(s)* (superlative). With suffixal plurals, which will require at most 20 additional graphs, the number of inflectional graph for Arabic nouns does not reach the number of graphs for the Unitex Portuguese (Brazil) dictionary.

### 7. Rules of agreement with broken plural nouns

The difference between BP and suffixal plural in Arabic is obviously a matter of inflectional morphology, but not only. Grammatical agreement of plural nouns with adjectives, participles or verbs is slightly different depending on whether the plural noun is a BP or a suffixal plural. The difference is observed both with human and non-human nouns, but agreement follows distinct rules.

#### 7.1. Human nouns

A human noun in the plural can agree with adjectives and participles in the broken or suffixal plural, or with both, if the adjective has both plurals. This rule applies independently of whether the plural noun is a BP, as *EulamaAocu* 'scientists', or a suffixal plural, as *muraAoqibuwona* 'observers'. In the following examples, the *:q* code marks BPs, and *:p* marks suffixal plurals:

....والعلماء (العاملون + النشطاء) في حقل الكيمياء ...
```
wa-Al-EulamaAcu      Al-(nuMaTaAc + EaAmiluwna) fiy Haqoli Al-kiymoyaAc
and-the-scientists:q the-(active:q + working:p) in  area   the-chemistry
```
'and the scientists (active + working) in the area of chemistry'

....و المراقبون الدوّليون (العاملون + النشطاء) في سوريا ...
```
wa-Al-muraAqibuwna  Al-duwGaliyGuna   Al-(nuMaTaAc + EaAmiluwna) fiy suwriyGaA
and-the-observers:p the-international the-(active:q + working:p) in  Syria
```
'and the international observers (active + working) in Syria'

However, if the human noun is in the BP, it can also agree with an adjective or participle in the feminine singular (*:fs* code below), no matter the gender of the noun or the sex of its referent:[22]

....والعلماء العاملة في حقل الكيمياء ...
```
wa-Al-EulamaAcu         Al-EaAmilapu   fiy Haqli Al-kiymoyaAc
and-the-scientists:mq   the-working:fs in  area  the-chemistry
```
'and the scientists working in the area of chemistry'

This additional possibility of agreement is not observed with suffixal plurals of human nouns (the '*' symbol signals unacceptability here):

*... والمراقبون الدوّليون العاملة في سوريا ...*
```
*wa-Al-muraAqibuwna  Al-duwGaliyGuna   Al-EaAmilapu   fiy suwriyGaA
*and-the-observers:p the-international the-working:fs in  Syria
```
'and the international observers working in Syria'

Agreement of adjectives in the feminine singular with BP human nouns may surprise non-Arabic speakers. It is less frequent than agreement of adjectives in the plural, but handbooks definitely consider it as grammatical, and it occurs in literary works:

[22] The adjective or participle could be analysed and labeled as an alternative plural, with the same form as a feminine singular (Smrž, 2007:27).

... الرجال شحيحة في مصر الآن.

*Al-rijaAlu MaHiHapun fiy misra AaloCn*

the-<N:mq> <A:fs> in Cairo presently

'Men are rare in Cairo presently'

(Rim Basyuwniy, *Smell of The Sea*, http://arabicorpus.byu.edu/)

The rules of grammatical agreement between subject noun and verb, when the verb occurs after the subject, are similar to the rules above. A BP human noun subject can agree with the verb in the feminine singular, whereas a suffixal plural human noun subject cannot:

القضاة (غادرت + غادروا + غادرن ) ظهراً

*Al-quDaApu (gaAdarat + gaAdaruwA + gaAdarona) ZuhoraAF*

The-judges:q (left:fs + left:mp + left:fp) at-mid-day

'The judges left at-mid-day'

المراقبون (*غادرت + غادروا + *غادرن ) ظهراً

*Al-muragibuwna (*gaAdarat + gaAdaruwA + *gaAdarona) ZuhoraAF*

The-observers:mp (*left:fs + left:mp + *left:fp) at-mid-day

'The observers left at mid-day'

### 7.2. Non-human nouns

With non-human nouns, agreement rules are slightly different, but they still discriminate between BPs and suffixal plurals. Both types of plural can agree with an adjective or participle in the feminine singular, but only suffixal plurals can agree with an adjective or participle in the plural (*:fp* code below):

إستعملت ( *المعاول + الحلقات ) الصالحات

*IistaEomaltu Al-(*maEaAwilu + HalaqaAtu) SaAliHaAtun*

I used the-(*mattocks:q + rings:fp) good:fp

'I used the good (mattocks + rings)'

A dozen non-human nouns with BP, often denoting female animals, are exceptions to this rule and can agree with an adjective or participle in the plural.

### 7.3. Codification

The formalization of agreement rules in parsers and generators requires discrimination between the BP and suffixal plural of Arabic nouns. We opted for the straightforward solution of distinguishing two values for number, *q* and *p*. Taking into account the singular and the dual, our morpho-syntactic model of Arabic totals 4 values for number of nouns and adjectives. The MAGEAD system (Altantawy *et al*., 2011) has 3 values for number: singular, dual and plural. The Smrz (2007) parser has 3 values also.

We lack bases to define the gender of a BP. Broken plural shows no morphological difference in gender, even when the singular does: *qaAoDiy* 'male judge' and *qaAoDiyap* 'female judge' have the same BP *quDaAop* 'male or female judges or both'. Rules of agreement of a human BP with adjectives in the suffixal plural: *<A:mp>*, *<A:fp>*, or with verbs in the plural, depends on the sex of the referent. In the case of a non-human BP, an agreeing adjective is obligatorily in the feminine singular. Thus, our model represent BPs without any gender, tagging them as *<N:q>*.

## 8. Clitic-related spelling variants

In Arabic, a token can be analysed as a sequence of segments. Each segment in a token is a morpheme. A nominal token may contain a single morpheme *<N>*, or the concatenation of up to 5 morphemes as in:

*<CONJC> <PREP> <DET><N> <PRO+Gen>*

where *<CONJC>* is a coordinating conjunction, <PREP> a preposition, *<DET>* the determiner *Al-*, and *<PRO+Gen>* a pronoun in the genitive. The combination of morphemes obeys a number of constraints. A *<PREP>* constrains the noun to be in the genitive case.[23] The presence of a clitic, graphically agglutinated *<PRO+Gen>* constrains another inflectional feature of the noun, definiteness, to have the construct-state value, while two other values, definite and indefinite, are possible otherwise. By checking such constraints, wrong segmentations can be discarded.

### 8.1. Segmentation

With the Unitex system, we represent nouns with four inflectional features: gender (masculine, feminine), number (singular, dual, suffixal plural, BP), definiteness (definite, indefinite, construct-state) and case (nominative, accusative, genitive). The segmentation into morphemes is performed with the aid of graphs. The output of this process is saved in the text automaton as in Fig. 1.

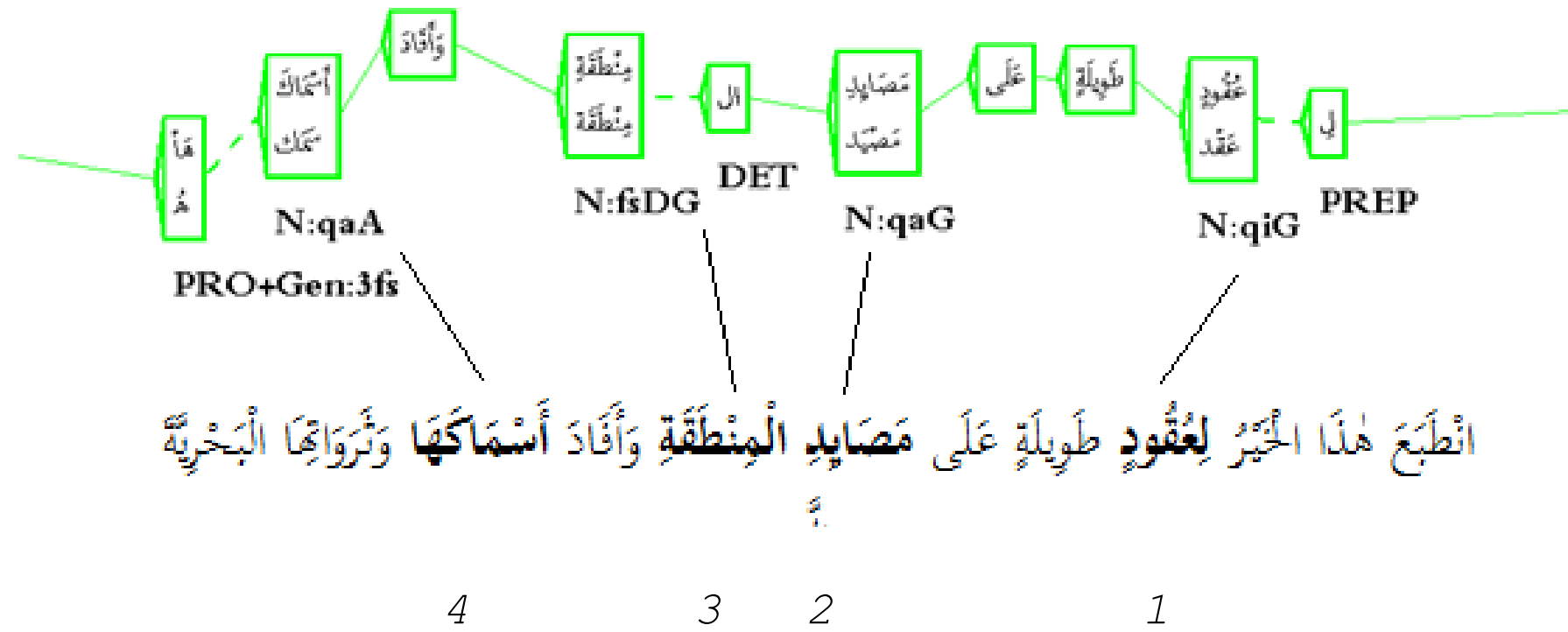

Fig. 1. Nouns tagged in text. Text automaton resulting from the application of graphs of morphological segmentation. Dashed lines connect segments inside the same token.

The sequence displayed in Fig. 1 contains 4 nouns, among which 3 BPs:

| No. | | Token | Lexical item |
|---|---|---|---|
| 1 | BP | *li_Euquwd-K* | *Eaqod,FvEvL-FuEuuL-123* |
| 2 | BP | *maSaAyid* | *maSoyad,FvEvLvB-FaEaaLiB-1234* |
| 3 | sing. | *Al_minoTaqap-i* | *minoTaqap,FvEvLvB-FaEaaLiB-1234* |
| | (This singular noun is labelled by the analyser since it admits a BP) | | |
| 4 | BP | *OasmaAk-i_haA* | *samak,FvEvL-OaFoEaaL-123* |

Dashed lines connect segments inside the same token. Abbreviations read as follows: PREP (preposition), DET (determiner), PRO (pronoun), Gen (genitive). Genders: **m**asculine, **f**eminine. Numbers: **s**ingular, **d**ual, suffixal **p**lural, **q** for broken plural. Definitenesses: **D**efinite, **i**ndefinite, and **a** for construct-state. Cases: **N**ominative, **A**ccusative, **G**enitive.

### 8.2. Orthographic adjustments

Most inflected noun forms are insensitive to graphically agglutinated pronouns, but some forms undergo an orthographic adjustment, e.g. forms with the suffix *-ap* or ending with a glottal stop. The suffix *-ap* is realised as its allograph *-at-*. In the full-form dictionary, those morphological variants that combine with the pronoun are marked as *<N+pro>*. Segmentation graphs select the *<N+pro>* variants from the dictionary. Fig. 2 shows the text automaton resulting from the morphological analysis of *OanoMiTatihaA* 'its activities':

| No. | | Token | Lexical item |
|---|---|---|---|
| 1 | BP | *OanoMiTaṯ-i-haA* | *naMaAT,FvEvvL-OaFoEiLap-123* |

[23] *<CONJC>* combines freely with any inflected noun.

The segmentation graph checks that the agglutinated variant is marked as <*N+pro*> in the dictionary. Dashed lines connect segments inside the same token.

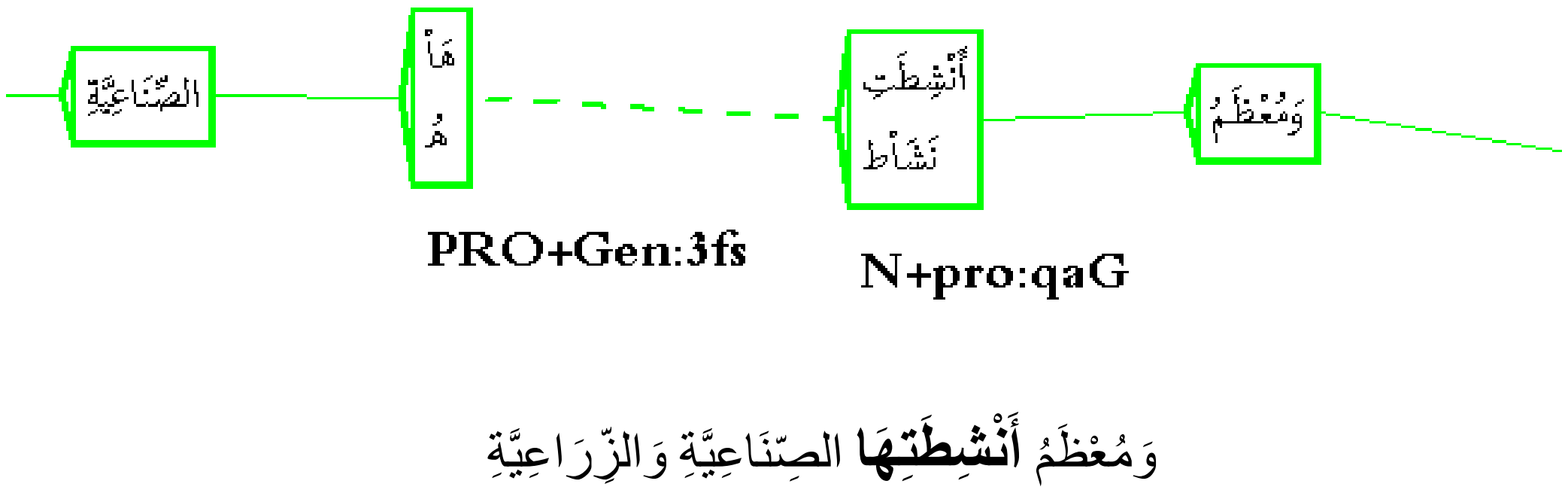

Fig. 2. Text automaton resulting from morphological segmentation.

The generation of the orthographically adjusted variants of an inflected noun is performed directly during the compilation of the dictionary of word forms. This process applies rules of orthographical variation, but makes use of lexical information encoded in entries. During analysis, the segmentation graph links each morphological variant to the correct context: again, this process implements rules, but takes advantage of formalized lexical information. The variants are generated during the compilation of the resources, not at analysis time as in rule-based systems in which a rule should compute each morphological variant at run time, then link each variant to the correct context. Our method simplifies and speeds up the process of annotation.

The system generates the inflected forms with the aid of an inflectional transducer (Fig. 3), as in Silberztein (1998). This transducer invokes sub-graphs; one of them, displayed in Fig. 4, specifies the generation of the orthographically adjusted construct-state variants (with the form -*at*- of the suffix) of an inflected form. The generation is performed during the compilation of the dictionary.

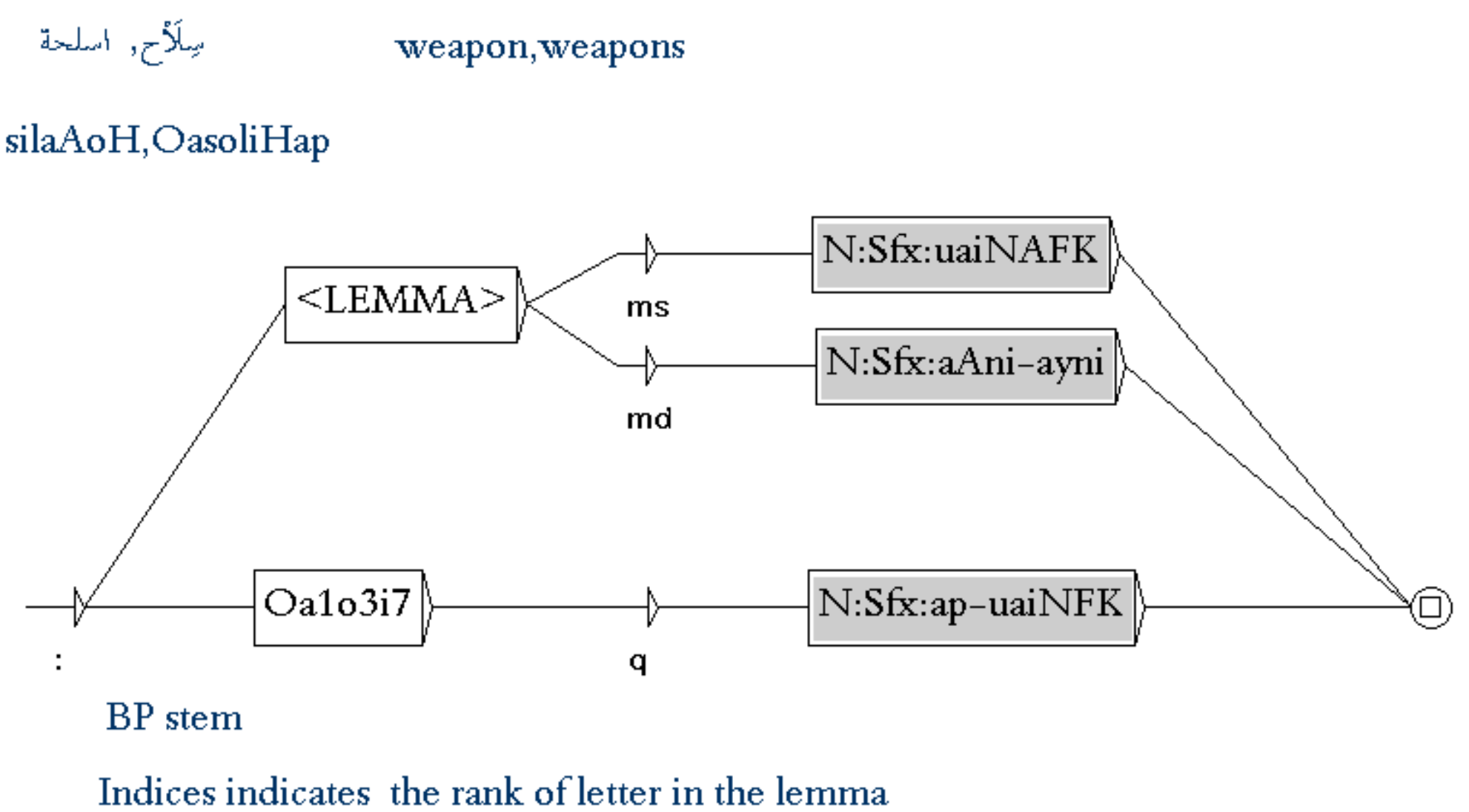

Fig 3. Inflectional transducer N300-m-FvEvvL-OaFoEiLap-123. Each path contains a stem pattern and a call to a subgraph of suffixes for definiteness and case variations (3×3).

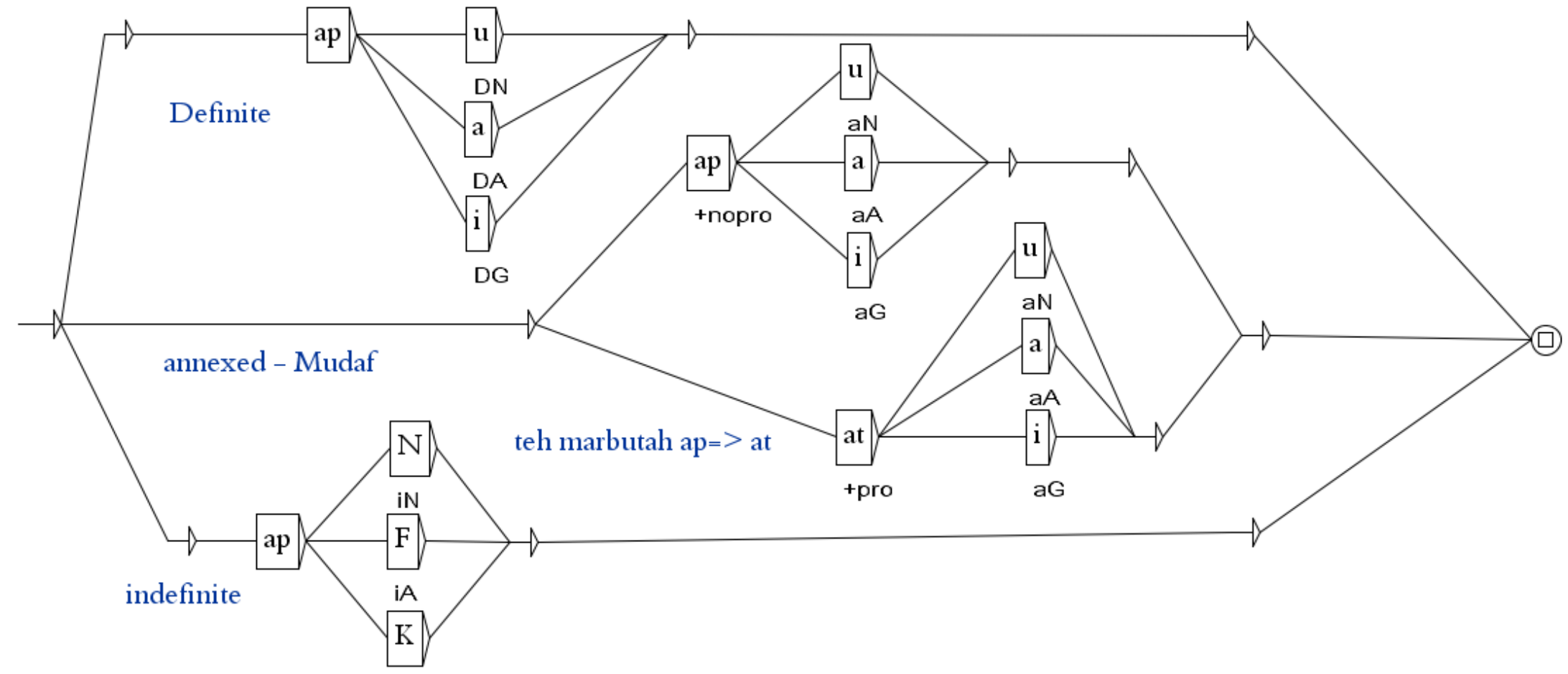

Fig. 4. Subgraph ap-uaiNFK represents definiteness/case suffix variations for nouns ending with the suffix *-ap*.

## 9. Evaluation

Since our BP lexicon is partial, we have chosen to measure its lexical coverage, and the feasibility of the extension of lexical coverage.

### 9.1. Corpus

We used a small sample of the NEMLAR Arabic Written Corpus (Attia *et al.*, 2005). This corpus was produced and annotated by RDI, Egypt, for the Nemlar Consortium.[24] During the construction of our lexicon of BPs, we did not use any part of the corpus: our sources of information were handbooks, reference dictionaries and native speaker competence. Thus, the evaluation tool is independent from the evaluated resource.

We selected three documents totalling 3 550 tokens (about 10 pages) and containing scientific popularization about three topics: pollution and fishing in Egypt, earthquakes in the world, and quality of water. We used the documents in the fully diacritized version.[25]

### 9.2. Coverage

We have extracted manually 388 occurrences of plural nouns and adjectives: 267 BPs and 121 suffixal plurals, among which 8 in the masculine and 113 in the feminine. Our lexicon (3 198 entries with BP) covered 195 occurrences out of the 267, i.e. 73% of occurrences. The sample did not contain any adjective in the BP.

The 195 covered occurrences of BPs are forms of 84 different lemmas of nouns, while the 72 remaining occurrences are forms of 25 lemmas of nouns: the lexicon covered 77% of the lemmas in the sample.

The 267 occurrences of BPs belong to 33 different inflectional classes, which had all been encoded in the system before evaluation. During the evaluation experiments, 5 descriptions of classes were found to contain errors affecting the recognition or tagging of forms. Therefore, the system covered 100% of the inflectional classes relevant for the sample, and 85% of them without errors.

| | Sample | Covered | Coverage |
|---|---|---|---|
| Occurrences | 267 | 195 | 73% |
| Lemmas | 109 | 84 | 77% |
| Inflectional classes | 33 | 33 | 100% |

[24] It consists of about 500 thousand words of Arabic text from 13 different genres. Each text is provided in 4 different versions: raw text, fully diacritized text, text with Arabic lexical analysis, and text with Arabic POS-tags.
[25] The annotated corpus (10 pages) will be freely available in a file named *Fishing-Earthquakes-Water.txt* in the Unitex/Arabic/Corpus folder.

BP occurrences make up 7.5% of all tokens of the sample, but 69% of all occurrences of plural nouns and adjectives, a surprisingly high proportion. In order to check this point, we made another study with another document from the Nemlar corpus, belonging to another genre: a 2 510-token biographical text (4 pages) by Tawfiq Hakim, an Egyptian playwright. We counted 158 BP occurrences, which make up 6.3% of all tokens, and 73% of the 216 plural nouns and adjectives.

Thus, in spite of the fact that BPs are irregular, their presence in Arabic text is predominant over suffixal plurals. To our knowledge, this quantitative predominance had not been discovered before.

Among the 267 BP occurrences, 170 occurrences (64%) are graphically agglutinated with other segments and 97 are not. This means that graphical agglutination affects nouns in a massive way.

### 9.3. Feasibility of the extension of lexical coverage

The 72 occurrences of BP missing in the lexicon were analysed as forms of 25 distinct lemmas, for which 25 new entries were inserted. All new entries were assigned to already encoded inflectional classes. The new entries were tested by compiling the lexicon and tagging the evaluation corpus. The description of one of the classes had to be corrected because of a filename error. The analysis, encoding, testing and correction required 4 hours' work.

This experiment validates the feasibility of a comprehensive BP lexicon on the basis of the PRIM model.

The following list is a part of a concordance of the 267 occurrences of BPs in the evaluation corpus. It has been produced after lexicon update, by submitting the *<N:q>* lexical mask to Unitex:

اَلنَّيِتْرُوجِينِيَّةِ وَالْفِسْفُورِيَّةِ وَالْمَوَادِّ الْعُضْوِيَّةِ الْأُخْرَي الَّتِي
عُضْوِيَّةِ الْأُخْرَي الَّتِي تُعْتَبَرُ غِذَاءً لِلْأَسْمَاكِ وَالَّتِي تَفِدُ إِلَي الْبَحْرِ
الْبَحْرِ وَالْبُحَيْرَاتِ مَعَ مِيَاهِ النِّيلِ وَالْمَصَارِفِ وَالْقَنَوَاتِ الزِّرَاعِيَّةِ
الْآدَمِيَّةِ الَّتِي تَذْهَبُ لِلْبَحْرِ أَمَامَ سَوَاحِلِ الدَّلْتَاْ وَمَصَايِدِهَا مُنْذُ بِدَاي
تَذْهَبُ لِلْبَحْرِ أَمَامَ سَوَاحِلِ الدَّلْتَاْ وَمَصَايِدِهَا مُنْذُ بِدَايَةِ الثَّمَانِينِيَّات
ذُ بِدَايَةِ الثَّمَانِينِيَّاتِ .وَهٰذِهِ الْأَسْبَابُ هِيَ :أَوَلًا :زِيَادَةُ التِّع
حُوظُ فِي حَجْمِ وَمِسَاحَةِ شَبَكَاتِ الْمِيَاهِ وَالْمَجَارِي وَمَحَطَّاتِ الصَّرْفِ الْآدَمِيِّ
وَمَحَطَّاتِ الصَّرْفِ الْآدَمِيِّ بِخَاصَّةٍ فِي مُدُنِ الْقَاهِرَةِ وَالْإِسْكَنْدَرِيَّةِ خِلَالَ
حَقِيقَةٌ أَمْ وَهْمٌ؟ !لَمْ يَتَّفِقْ بَعْضُ الْعُلَمَاءِ مَعَ رَأْيِ نِكْسُونْ وَمِنْهُمُ
لْمِصْرِيُّ الدُّكْتُورُ يُوسُفُ حَلِيمٍ بِقِسْمِ عُلُومِ الْبِحَارِ بِجَامِعَةِ الْإِسْكَنْدَرِيَّة
يُّ الدُّكْتُورُ يُوسُفُ حَلِيمٍ بِقِسْمِ عُلُومِ الْبِحَارِ بِجَامِعَةِ الْإِسْكَنْدَرِيَّةِ، حَيْث
فِي الْمَقَامِ الْأَوَّلِ إِلَى زِيَادَةِ عَدَدِ سُفُنِ وَمَرَاكِبِ الصَّيْدِ وَتَحَسُّنِ كَفَاءَةِ
مَقَامِ الْأَوَّلِ إِلَى زِيَادَةِ عَدَدِ سُفُنِ وَمَرَاكِبِ الصَّيْدِ وَتَحَسُّنِ كَفَاءَةِ الْمُعَد
الصَّيْدِ وَتَحَسُّنِ كَفَاءَةِ الْمُعَدَّاتِ وَالْأَجْهِزَةِ الْمُسْتَخْدَمَةِ فِي عَمَلِيَّاتِ
زِيَادَةِ نِسْبَةِ الْهَائِمَاتِ النَّبَاتِيَّةِ وَالْمَوَادِّ الْعُضْوِيَّةِ .كَمَا يُؤْخَذُ
مْلَةِ الْمُلَوِّثَاتِ الَّتِي ذَكَرَتْهَا بَيْنَ الْمَوَادِّ الْعُضْوِيَّةِ الْمَوْجُودَةِ فِي الْأ
الْمَوَادِّ الْعُضْوِيَّةِ الْمَوْجُودَةِ فِي الْأَسْمِدَةِ وَالْمُخَلَّفَاتِ اَلْآدَمِيَّةِ وَب

In order to investigate the feasibility of the extension of lexical coverage beyond BPs and verbs (Neme, 2011), we inserted in the lexicon 750 items for all the words occurring in the evaluation corpus and not found in the lexicon. We encoded 52 inflectional classes for suffixal plural nouns, suffixal plural adjectives, grammatical words and for 2 classes of verbs missing in Neme (2011). The encoding and the testing/correction loop required 60 hours' work. After this extension, the evaluation corpus was entirely covered.

This experiment validated our intuition that, besides verb conjugation and BPs, Arabic morpho-syntactic tagging does not pose any serious challenges to resource-based language processing.

## Conclusion

By keeping inflection apart from derivational morphology and dealing with morphophonological alternations in a factual way, the PRIM model simplifies the encoding of BP. Its strong points can be summed up as follows:

1. It complies with the conventions in traditional morphology that we found useful to noun inflection, in particular with most of the traditional patterns in the sense of Semitic morphology. Thus, the PRIM language resources can be easily updated by Arabic-speaking linguists in order to extend lexical coverage and control the evolution of the accuracy of systems that use them. We have dropped conventions related to semantic description.

2. The updatable lexicon is structured in lexical entries, as traditional dictionaries, and not in stem entries, as in the multi-stem approach.

3. Inflected forms are generated from their observable surface lemma, and not from a deep root.

4. The pattern of a singular noun is abstracted from the stem without gender or number suffixes, and without definiteness and case markers. The pattern of a BP is abstracted from the stem without definiteness or case markers.

5. The taxonomy of singular patterns specifies vowel quantity, noted as *v* or *vv*, but ignores vowel quality and derivational history.

6. Patterns are not used to represent morpho-syntactic features in lexical tags. Lexical tags are accurate and informative and consist of a lemma and a set of feature-value pairs, generally gender, number, definiteness and case.

7. Root alternations are encoded independently from patterns. They are explicitly represented as separate pieces of lexical information, instead of being obtained through the interaction of a deep level with general rules. They are encoded as mappings from the surface root of the singular to the surface root of the plural. Orthographical variations of the glottal stop are encoded in the same way.

8. Root letter substitutions and insertions are restricted to *w*, *y*, *A*, to allographs of the glottal stop, and to copies of root letters available in the lemma.

9. The PRIM taxonomy for noun inflection is simple, orderly and detailed. The number of classes, including suffixal plural and BP, is smaller than for Brazilian Portuguese.

10. A transducer corresponds to each inflectional class of nouns, and generates all the inflected forms of any lemma in the class. Transducers are edited in graphical form with the Unitex system, and handle roots in Semitic languages straightforwardly. They can be quickly corrected when an error is detected.

11. Morphological analysis of Arabic text is performed directly with a dictionary of words and without morphological rules, which simplifies and speeds up the process.

12. Agglutinated clitics are analysed without generation of artificial ambiguity. Clitic agglutination is described independently from inflection, in separate graphs.

13. The PRIM model is compatible with solutions to the other challenges to Arabic processing: verb conjugations, including alternations of *w*, *y*, *A* and the glottal stop (Neme, 2011); recognition of partially diacritized text with fully diacritized resources, excluding incompatible analyses.

Our distinctive approach consists in considering language resources as the key point of the problem. We integrate all complex operations among resource management operations.

## Bibliography

Abdel-Nour, Jabbour (2006). *Dictionnaire Abdel-Nour al-Mufassal Arabe-Français*. Dar El-Ilm Lil-Malayin. 10th edition. 2034 pages, 3 columns.

Altantawy, Mohamed; Habash, Nizar; Rambow, Owen (2011). Fast Yet Rich Morphological Analysis. In *Proceedings of the 9th International Workshop on Finite State Methods and Natural Language Processing* (FSMNLP), pages 116-124.

Altantawy, Mohamed; Habash, Nizar; Rambow, Owen; Saleh, Ibrahim (2010). Morphological Analysis and Generation of Arabic Nouns: A Morphemic Functional Approach. In *Proceedings of the Language Resource and Evaluation Conference* (LREC), Malta, pages 851-858.

Attia., M., Yaseen., M., Choukri., K. (2005). *Specifications of the Arabic Written Corpus produced within the NEMLAR project*, www.NEMLAR.org.

Beesley, Kenneth R. (1996). Arabic finite state morphological analysis and generation. In *Proceedings of the International Conference on Computational Linguistics* (COLING), Copenhagen, Center for Sprogteknologi, volume 1, pages 89-94.

Beesley, Kenneth R. (2001). Finite-State Morphological Analysis and Generation of Arabic at Xerox Research: Status and Plans in 2001. In *Proceedings of the ACL/EACL Workshop 'Arabic Language Processing: Status and Prospects'*, pages 1-8.

Boudlal, Abderrahim; Lakhouaja, Abdelhak; Mazroui, Azzeddine; Meziane, Abdelouafi (2010). Alkhalil Morpho SYS1: A Morphosyntactic Analysis System for Arabic Texts. *International Arab Conference on Information Technology* (ACIT).

Brame, M. (1970). *Arabic Phonology: Implications for Phonological Theory and Historical Semitic*, unpublished Ph.D. dissertation, MIT.

Buckwalter, Timothy (1990). Lexicographic notation of Arabic noun pattern morphemes and their inflectional features. In *Proceedings of the Second Cambridge Conference on Bilingual Computing of Arabic and English*. 7 pages.

Buckwalter, Timothy. *Arabic Morphological Analyzer* Version 1.0. (2002). LDC Catalog No.: LDC2002349.

Buckwalter, Timothy (2007). Issues in Arabic Morphological Analysis. In Antal van den Bosch and Abdelhadi Soudi (eds.), *Arabic Computational Morphology*. *Knowledge-based and Empirical Methods*. Text, Speech and Language Technology, volume 38, Berlin: Springer, pages 23-41.

Courtois, Blandine (1990). Un système de dictionnaires électroniques pour les mots simples du français, *Langue Française* 87, Paris: Larousse, p.11-22.

El-Dahdah Antoine (1992), *A dictionary of universal Arabic Grammar*. Librairie du Liban Publishers. Bilingual, 250 p. in Arabic/250 p. in English.

Daille, Béatrice; Fabre, Cécile; Sébillot, Pascale (2002). Applications of Computational Morphology. In Boucher, Paul (ed.), *Many Morphologies*, Somerville: Cascadilla Press, p. 210-234.

Al-Fairuzabadi (v. 1400), *Al-Qamus al-Muhit* (Comprehensive Dictionary). Ed. 2007, Beirut: Dar Al Kotob Al Ilmiyah, 1440 pages.

Ferrando, Ignacio (2006). The plural of paucity in Arabic and its actual scope. On two claims by Siibawayhi and al-Farraa'. In: Boudelaa, Sami (ed.), *Perspectives on Arabic Linguistics*, XVI, Current Issues in Linguistic Theory, 266, Amsterdam/Philadelphia: Benjamins, p. 39-61.

Al-Ghalāyini, Mustafa (2007). *Jāmi3 al-durūs al-'arabiyah* (A university grammar textbook). 1st edition 1912. Dar El Fikr Printers-Publishers, Beirut. 570 pages. In Arabic.

Gross, Maurice (1975). *Méthodes en syntaxe. Régime des constructions complétives*. Paris: Hermann.

Haaruun, S.M. (ed.) (1977). 2ª. Sibawayh (around 800 CE), Kitaabu Siibawayhi 'Abii Bišrin 'Amri bni 'Utmaana bni Qunbur, Cairo, 5 vols.

Habash, Nizar; Rambow, Owen (2006). MAGEAD: A Morphological Analyzer and Generator for the Arabic Dialects. *In Proceedings of the International Conference on Computational Linguistics and Annual Meeting of the Association for Computational Linguistics* (COLING-ACL), Sydney, Australia, pages 681–688.

Al-Hadithy, Khadija (2003). *Morphological forms in the Sibawayh's Kitāb. A dictionary and a study*. Republication of the dissertation in the Master of Arts, Faculty of Literature in Cairo, first published in 1961. 370 pages. In Arabic.

Huh, Hyun-Gue; Laporte, Éric (2005). A resource-based Korean morphological annotation system. In *Proceedings of the International Joint Conference on Natural Language Processing* (IJCNLP), Jeju, Korea.

Ibn Manzur (1290). *Lisān al-ʿArab* (The Arabic Language). Ed. 1955-1956, Beirut: Dar Sadir, 15 volumes.

Kihm, Alain (2006). Nonsegmental concatenation : a study of Classical Arabic broken plurals and verbal nouns, *Morphology* 16, 69-105.

Kiraz, George Anton (1994). Multi-tape Two-level Morphology: A Case study in Semitic Non-Linear Morphology. In *Proceedings of the International Conference on Computational Linguistics* (COLING), Kyoto, Japan, pages 180–186.

Kiraz, George Anton (2001). *Computational Nonlinear Morphology, with Emphasis on Semitic Languages* Cambridge, U.K.: Cambridge University Press. Studies in Natural Language Processing, 171 pages.

Lane, Edward William (1893). *Arabic-English Lexicon*. Williams and Norgate, London.

Levy, M. M. (1971). *The plural of the nouns in Modern Standard Arabic*, PhD dissertation, University of Michigan.

McCarthy, J. J. (1981). A prosodic theory of nonconcatenative morphology. *Linguistic Inquiry* 12, 373-418.

McCarthy, J.J. (1983). *A prosodic account of Arabic broken plurals*. Linguistics Department Faculty Publication Series, Paper 25, University of Massachusetts – Amherst, pp. 289–320.

McCarthy, J. J. & Prince, A. S. (1990). Foot and word in prosodic morphology: the Arabic broken plural. *Natural Language and Linguistic Theory* 8(2), 209–283.

Muniz, Marcelo C.M.; Maria das Graças V. Nunes; Éric Laporte (2005). UNITEX-PB, a set of flexible language resources for Brazilian Portuguese. In *Proceedings of the TIL Workshop*. pages 2059–2068.

Murtonen, Aimo (1964). *Broken Plurals: Origin and Development of the System*. E.J, Brill, Leiden.

Neme, Alexis (2011) A lexicon of Arabic verbs constructed on the basis of Semitic taxonomy and using finite-state transducers. In *Proceedings of the International Workshop on Lexical Resources* (WoLeR) at ESSLLI.

Paumier, Sébastien. (2011). *Unitex - manuel d'utilisation 2.1* , Université Paris-Est Marne-la-Vallée.

Ratcliffe, Robert R. (1998), *The 'broken' plural problem in Arabic and comparative Semitic: allomorphy and analogy in non-concatenative morphology*, John Benjamins, Foreign Language Study, 261 pages.

Ratcliffe, Robert R. (2001), Analogy in Semitic morphology: where do new roots and new patterns come from? In Zaborsky, Andrzej (ed.), *New Data and New Methods in Afroasiatic Linguistics. Robert Hetzron in memoriam*, Wiesbaden, Harrassowitz, p. 153-162.

Ryding C. Karin (2005), *A Reference Grammar Of Modern Standard Arabic*, Cambridge University Press, 708 pages.

Silberztein, Max (1998). INTEX: An integrated FST toolbox. In Derick Wood, Sheng Yu (eds.), *Automata Implementation*, p. 185-197, Lecture Notes in Computer Science, vol. 1436. Second International Workshop on Implementing Automata, Berlin/Heidelberg: Springer.

Smrž, Otakar (2007). *Functional Arabic Morphology. Formal System and Implementation*. Ph.D. thesis, Charles University in Prague, Czech Republic.

Soudi, Abdelhadi; Cavalli-Sforza, Violetta; Jamari, Abderrahim (2002), The Arabic Noun System Generation.

Tarabay, Adma (2003). *A dictionary of Arabic plurals.* Librairie du Liban Publishers. 590 pages. In Arabic.

Wehr, Hans (1960). *Dictionary of Modern Written Arabic*. Spoken Language Services, Ithaca, N.Y.

Wright, William (1971). *A Grammar of the Arabic Language*. Cambridge University Press.

Zbib, Rabih; Soudi, Abdelhadi (2012). Introduction. In Soudi, Abdelhadi; Farghaly, Ali; Neumann, Günter; Zbib, Rabih (eds.), *Challenges for Arabic Machine Translation*, Natural Language Processing, 9, Amsterdam: Benjamins, p. 1-13.